%% file: main.tex
\documentclass{article}
\pdfoutput=1

\usepackage[main,final,nonatbib]{neurips_2026}

\usepackage[utf8]{inputenc}
\usepackage[T1]{fontenc}
\usepackage{amsmath,amssymb,mathtools}
\usepackage{booktabs}
\usepackage{array}
\usepackage{tabularx}  %
\usepackage{longtable}   %
\usepackage{algorithm}
\usepackage{algorithmic}
\usepackage{placeins} %
\usepackage{graphicx}
\usepackage{subcaption}
\usepackage{xcolor}
\usepackage[numbers,square,sort&compress]{natbib}
\setcitestyle{numbers,square,comma,sort&compress}
\PassOptionsToPackage{hyphens}{url}     %
\usepackage{hyperref}
\hypersetup{pdfencoding=auto,
  pdfauthor={Michal Pr\r{u}\v{s}ek, Adam Novoz\'amsk\'y, Filip \v{S}roubek}}
\makeatletter
\AtBeginDocument{\hypersetup{pdftitle={\@title}}}
\makeatother

\newcommand{\geowalk}{\textsc{GeoWalk}}
\newcommand{\geocma}{\textsc{GeoCMA}}
\newcommand{\R}{\mathbb{R}}

\newcommand{\ip}[2]{\langle #1,#2\rangle}
\newcommand{\norm}[1]{\lVert #1 \rVert}
\newcommand{\pb}[1]{\textbf{#1}}

\newif\ifcrprelim
\input{cr_status.tex}

\input{numbers.tex}

\newcommand{\crRepoUrl}{\url{https://github.com/michalprusek/geometry-before-surrogates-neurips2026}}
\newcommand{\crProjectUrl}{\url{https://staff.utia.cas.cz/novozada/geo}}
\ifcrprelim
  \usepackage{eso-pic}
  \AddToShipoutPictureBG{\AtPageUpperLeft{\raisebox{-0.6in}{\makebox[\paperwidth]{%
    \color{red}\sffamily\bfseries\small PRELIMINARY NUMBERS (legacy data) --- generated tables,
    figures and numbers are placeholders, not the camera-ready values}}}}
  \DeclareCaptionLabelFormat{crprelim}{\leavevmode\llap{\textcolor{red}{%
    \sffamily\bfseries\scriptsize LEGACY\enspace}}\bothIfFirst{#1}{~}#2}
\fi
\newcommand{\crinputifexists}[1]{%
  \IfFileExists{#1}{\input{#1}}{%
    \PackageWarningNoLine{crkit}{Generated file #1 not found (written by the production run only)}%
    \ifcrprelim\par\noindent{\color{red}\fbox{\parbox{0.96\linewidth}{\footnotesize\sffamily
      \texttt{\detokenize{#1}}: this generated table does not exist in the legacy output; it appears
      here once the production files are copied in.}}}\par\fi}}
\makeatletter
\newcommand{\crtableplacement}[1]{%
  \renewenvironment{table}[1][]{%
    \setlength{\abovecaptionskip}{\@neuripsbelowcaptionskip}%
    \setlength{\belowcaptionskip}{\@neuripsabovecaptionskip}%
    \@float{table}[#1]}{\end@float}}
\makeatother

\title{Match the Geometry Before the Surrogate: Extreme Low-Budget Optimization in High Dimensions}

\author{%
  Michal Pr\r{u}\v{s}ek$^{1,2}$ \qquad Adam Novoz\'amsk\'y$^{2}$ \qquad Filip \v{S}roubek$^{2}$ \\
  $^{1}$Faculty of Nuclear Sciences and Physical Engineering, Czech Technical University in Prague \\
  $^{2}$Institute of Information Theory and Automation, Czech Academy of Sciences, Prague \\
  \texttt{\{prusek,novozamsky,sroubekf\}@utia.cas.cz} \\
}

\begin{document}
\maketitle

\begin{abstract}
Low-budget molecular optimization is often framed as Bayesian optimization in a learned latent space, but standard benchmarks typically allow hundreds or thousands of evaluations, whereas a realistic wet-lab round may allow only a few dozen. We revisit this regime using a SELFIES molecular variational autoencoder (VAE) with a Gaussian prior and GuacaMol objectives at $N=30$ evaluations. The main finding is not a new surrogate model, but an interaction in candidate generation. The encoder means of a reference set of molecules concentrate near a narrow spherical shell, while many high-dimensional optimizers propose candidates from boxes or box-shaped trust regions. We study a simple constraint, the \emph{on-sphere scaffold}, that restricts candidate generation to this shell. We then introduce two minimal optimizers to probe what remains useful once the support is fixed. GeoWalk performs geodesic local search on the shell without a surrogate. After $30$ evaluations, it shows no detected difference from a Gaussian-process (GP) surrogate on the shell, which is ahead earlier in the budget. GeoCMA adds covariance learning in a random subspace. It does not improve on GeoWalk in the molecular benchmark. On LassoBench it has a lower mean loss than GeoWalk, but full-dimensional CMA-ES remains stronger on most tasks. A factorial experiment with a single GP crosses the candidate support (shell or box) with the acquisition maximizer (a finite candidate set or a continuous optimizer). In this Gaussian-prior VAE benchmark, shell candidates outperformed box candidates when acquisition was maximized over a candidate set, but not with continuous maximization by L-BFGS-B. These results come from a controlled study of one Gaussian-prior VAE benchmark and do not recommend VAEs for molecular optimization. The practical lesson is to control the candidate support and the acquisition maximizer before attributing a gain to the surrogate.
\end{abstract}

\section{Introduction}
\label{sec:intro}

Molecular and protein-design loops are often constrained by assays rather than by computation. A single wet-lab round may evaluate only a few dozen variants, each requiring substantial cost and weeks of turnaround time. This is far from the usual evaluation setting for high-dimensional Bayesian optimization (HD-BO)~\citep{eriksson2019turbo,maus2022lolbo,chu2024invbo}, where benchmark budgets often range from hundreds to thousands of oracle calls. In principle, Bayesian optimization is well suited to such expensive black-box objectives: it fits a cheaper surrogate model, typically a Gaussian process (GP) or another learned predictor, and uses it to choose the next candidate to evaluate. However, the budget mismatch raises a simple yet consequential question: \emph{In the extreme low-budget scenario ($30$ evaluations), what should the optimizer actually do?}

We study this question in a widely used molecular-design pipeline built around a SELFIES~\citep{krenn2020selfies} Transformer VAE, released with local latent space Bayesian optimization (LOL-BO)~\citep{maus2022lolbo} and pre-trained by its authors on the GuacaMol training set. The VAE has latent dimension $D=256$, and we evaluate it on the GuacaMol goal-directed tasks~\citep{brown2019guacamol}. This benchmark underlies several recent latent-space BO comparisons, and keeping its generator fixed lets us isolate the optimization layer. When the molecules from the ZINC-250k reference set~\citep{irwin2012zinc,gomez2018automatic} are encoded by the VAE, their encoder means concentrate near a narrow spherical shell rather than filling the surrounding box. Many HD-BO methods, however, generate candidates inside axis-aligned boxes or box-shaped trust regions. A Bayesian optimizer picks its next point by maximizing an acquisition function, a score of how promising a point is under the surrogate. It can score a finite set of randomly drawn candidates and take the best one, or it can start from a point and climb the acquisition function with a gradient-based optimizer, which we call continuous maximization. Here that optimizer is L-BFGS-B on both supports. In this Gaussian-prior VAE benchmark, shell candidates outperformed box candidates when acquisition was maximized over a candidate set, but not with continuous maximization by L-BFGS-B. The GP is fitted in the same way on both supports, but box candidates are rescaled per coordinate to $[0,1]^D$ before the GP, and shell candidates are not. With candidate-set maximization, the shell is ahead on \crFactorialSupportCandTasksK{} of the 11 tasks of our protocol.

The main object of study is therefore not a new surrogate model, but candidate generation: where candidates come from, and how an acquisition function is maximized over them. Instead of proposing candidates from an axis-aligned box, the \emph{on-sphere scaffold} places them on the empirical spherical shell occupied by the encoder means of the reference set. Formally, candidates have the form $x=\bar r u$, where $\bar r$ is the mean encoder norm of the run's cold-start molecules and $u\in S^{D-1}$ is a unit direction in the $D$-dimensional latent space. This constrains only the norm of a candidate, not its direction. We introduce two deliberately simple optimizers, \textbf{\geowalk{}} and \textbf{\geocma{}}, as probes of what remains useful once candidate generation is restricted to the encoder-mean shell. The first, \geowalk{}, samples the shell uniformly and then takes geodesic steps from the best point found so far. Despite having no surrogate, no learned covariance, and no acquisition function, \geowalk{} shows no detected difference from a GP on the shell after thirty evaluations, but the GP is ahead earlier in the budget. The second optimizer, \geocma{}, is positioned not as a competitor to \geowalk{} on the molecular VAE benchmark but as a complementary regime probe. It keeps the same on-sphere scaffold but replaces both phases of \geowalk{} with covariance learning in a random low-dimensional subspace, started from the best cold-start point, to mark where adding such structure becomes useful. On the molecular VAE benchmark itself \geocma{} does not improve over \geowalk{}. On two benchmarks outside chemistry, LassoBench~\citep{sehic2022lassobench} and NAS-Bench-101~\citep{ying2019nasbench}, it has better means than \geowalk{}. Full-dimensional CMA-ES, however, has the lower loss on most LassoBench tasks, and on NAS-Bench-101 \geocma{} is not detectably ahead of random search.

The paper makes three contributions. First, a factorial experiment shows that, in this benchmark, the advantage of shell candidates over box candidates depends on how the acquisition is maximized. Second, we introduce \geowalk{} and \geocma{} as minimal diagnostic optimizers that show what remains of this advantage without a GP. Third, all baselines are compared on one matched set, with every known deviation from the published methods reported (Appendix~\ref{app:baselines} and the method cards in the repository), because comparisons between latent optimizers can be confounded by candidate generation.

\textbf{Practical lesson.} When a learned latent space is optimized with few evaluations, first check where the candidates lie, for example by comparing their norms with those of encoded data, and how the acquisition is maximized. Otherwise, a gain attributed to the surrogate may come from candidate generation instead. A search without a surrogate on the same support, such as \geowalk{} in this paper, shows how much the surrogate adds.
\section{Related Work}
\label{sec:related}

Latent-space molecular optimization often follows a VAE-based pipeline: a variational autoencoder maps discrete molecules to continuous latent vectors, the latent space is searched with a Gaussian-process model of the objective, and selected latent points are decoded into molecules for scoring~\citep{gomez2018automatic}. Later methods refined different parts of this pipeline. LOL-BO~\citep{maus2022lolbo} introduced local trust regions in latent space. CoBO~\citep{lee2023cobo} regularized the VAE so that nearby latents have similar objective values. InvBO~\citep{chu2024invbo} addressed the encoder--decoder misalignment by inverting the decoder so that latent codes reconstruct their inputs consistently. NF-BO~\citep{lee2025nfbo} replaced the VAE altogether with an autoregressive normalizing flow that admits an exact left-inverse and removes the reconstruction gap. EULBO~\citep{maus2024eulbo} jointly optimizes the ELBO of its variational GP surrogate and a utility-based acquisition objective so that the surrogate is fitted with optimization in mind. COWBOYS~\citep{moss2025cowboys} decouples the GP surrogate from the generative model entirely, fitting them separately and combining them via a Bayesian update. The authors of COWBOYS also observed that molecular VAE latents occupy a thin shell, and argued that box-shaped search spaces cannot target it. These methods form the natural background for our study, but most of them are evaluated with budgets far larger than those available in a wet-lab round. We focus on the opposite regime: \emph{What remains reliable when the entire optimization loop consists of only a few dozen decoded and scored candidates?}

High-dimensional BO provides several mechanisms that could, in principle, help in this low-budget regime. TuRBO~\citep{eriksson2019turbo} uses local trust regions, BAxUS~\citep{papenmeier2022baxus} adapts nested low-dimensional embeddings, SAASBO~\citep{eriksson2021saasbo} introduces a sparse axis-aligned subspace prior on the GP lengthscales, and REMBO~\citep{wang2016rembo} and HeSBO~\citep{nayebi2019hesbo} search through random embeddings. A related linear-embedding line~\citep{letham2020alebo} uses Mahalanobis-style kernels on random subspaces; like \geocma{}, it restricts search to a low-dimensional subspace, but remains box-supported in the ambient space. Covariance Matrix Adaptation Evolution Strategy (CMA-ES)~\citep{hansen2001cmaes,hansen2019pycma} provides a complementary evolutionary baseline that does not rely on a probabilistic surrogate.

Linear BO~\citep{doumont2026still} applies an inverse stereographic projection inside the kernel of a linear GP; this changes the geometry seen by the model, while the search domain stays a cube. The on-sphere scaffold instead restricts where candidates are proposed. A complementary line of work designs Gaussian processes that respect a Riemannian manifold structure. GaBO~\citep{jaquier2020gabo} originally proposed Bayesian optimization on Riemannian manifolds, including the sphere, using a geodesic squared-exponential kernel. The Riemannian Mat\'ern variant~\citep{jaquier2022geometry} uses Mat\'ern Gaussian processes on Riemannian manifolds~\citep{borovitskiy2020matern}.

Recent work has also shown that Gaussian processes can remain competitive in high dimensions when their priors, lengthscales, and kernel defaults are chosen carefully~\citep{hvarfner2024vanilla,papenmeier2025understanding,xu2025standardgp}. These methods provide a strong basis for comparison. Our goal is not to show that Gaussian-process optimization cannot work in high dimensions, but to test how its results depend on where candidates come from and on how the acquisition is maximized. This view is also consistent with recent work emphasizing strong non-BO baselines~\citep{tripp2023genetic,gonzalezduque2024hdbo_survey}: in the VAE studied here, the appropriate random baseline is not arbitrary sampling from a box, but sampling from the encoder-mean shell. We note that direct discrete-space methods such as graph genetic algorithms~\citep{tripp2023genetic} bypass the VAE entirely and are highly competitive at larger budgets. We restrict scope to the question of how to use a learned latent space when one is already available, since this is the setting in which a candidate support must be chosen. We make no claim about which molecular model to use.
\section{Methods}
\label{sec:methods}

\subsection{Molecular optimization setup}
\label{sec:setup}
The molecular objectives are the $12$-task subset of the GuacaMol goal-directed benchmark~\citep{brown2019guacamol}, combining multi-property optimization (amlodipine, fexofenadine, osimertinib, perindopril, ranolazine, sitagliptin, zaleplon), deco hop, scaffold hop, valsartan SMARTS, and median-molecule (median molecule~1, median molecule~2) objectives. We adopt this $12$-task subset directly from the LOL-BO code~\citep{maus2022lolbocode}. The valsartan SMARTS task requires generating a molecule that matches a strict SMARTS pattern. No method tested produced a match at $N=30$, so the per-task score is $0$ for every row. We exclude it from cross-task aggregation rather than letting an all-zero column dominate noise statistics, leaving $11$ effective tasks. Because every row scores $0$ on it, leaving it out multiplies every cross-task mean by $12/11$ and changes no comparison. Cross-task means are means of per-task means. On a matched set with the same number of seeds in every task this equals the pooled mean, and the two differ only where a task has missing cells. A cell is one run of a method on one task with one seed. Each GuacaMol table, or block of a table, uses one matched set of cells, those completed by every row, and prints its size. A row that covers only part of that set is marked, and it uses the part it covers. The LassoBench tables print the coverage of every cell instead. Cells in which a failed GP fit forced a fallback step are included, and the repository counts them per method.

Each run is limited to $N=30$ evaluations. The first $5$ evaluations are fixed cold-start molecules shared by all optimizers for the same seed, and the optimizer then makes exactly $25$ proposals. A proposal that decodes to an already scored molecule, or does not decode, uses one of the $25$. The cold-start set is drawn without replacement from a fixed file of the first $10{,}000$ ZINC-250k molecules using the seed as random-number-generator state, so all methods see identical cold-start molecules within a seed but different molecules across seeds. They are the same for every task and method, are encoded by the encoder mean, and are scored directly. SELFIES tokens outside the VAE's vocabulary (stereo tokens) are removed before encoding; this affects $\crGeomTokenLossPct\,\%$ of the ZINC-250k molecules and every method alike. We report the best objective value found within these $30$ evaluations. For the main benchmark, each method is run on the $11$ non-degenerate GuacaMol tasks and on $40$ random seeds, numbered $42$--$81$. The runs of NAS-Bench-101, of two further codecs and with $N=1000$ (Section~\ref{sec:structured} and Appendices~\ref{app:crosscodec} and~\ref{app:budget}) are stored runs from an earlier code version and were not rerun. The BAxUS row of Table~\ref{tab:main} has $439$ of the $440$ cells, because the GP fit failed in one cell of the Zaleplon task.

The primary test of every comparison is the exact two-sided Wilcoxon signed-rank test~\citep{wilcoxon1945individual} on the per-task mean differences ($11$ tasks on GuacaMol, six on LassoBench), reported with the number of tasks on which a method is ahead. NAS-Bench-101 has a single task, so it is tested over seeds. The task is the unit because the cold-start molecules of a seed are shared across tasks, so the cells of one seed are not independent. Cell-level Wilcoxon tests are secondary, and so is Cohen's $d$, which we report as a cell-level effect size, computed as the mean paired difference divided by the standard deviation of paired differences. Benjamini--Hochberg correction~\citep{benjamini1995fdr} is applied within families: the five contrasts of the factorial experiment, the three recipe rows of Table~\ref{tab:boxsphere}, the tests of the published recipes (Table~\ref{tab:published}), and the LassoBench tests of the published methods. The secondary cell-level tests of the appendix tables are corrected within each block. The per-task comparison of \geowalk{} with the GP on the shell (Appendix~\ref{app:pertask}) uses Holm correction over the 11 tasks. The task-level $p$ in Tables~\ref{tab:main} and~\ref{tab:crossdomain} is uncorrected, and we call a result significant only within a corrected family.

The GuacaMol tasks have very different absolute score ranges (per-task means range from below $0.1$ to above $0.7$, Table~\ref{tab:pertask_full}). The standard error of the unweighted cross-task mean is therefore dominated by between-task heterogeneity rather than by between-method differences and is not informative for method comparison. We therefore base statistical claims on \emph{per-task} differences and do not attach standard-error bands to cross-task means.

\subsection{Empirical latent geometry and the \emph{on-sphere scaffold}}
\label{sec:scaffold}
The motivation for the on-sphere scaffold comes from the empirical geometry of the VAE latent space. When the $\crGeomNZinc$ molecules of the ZINC-250k reference set, which is not the VAE's training set, are encoded by the VAE, their encoder mean vectors $\mu(x)\in\R^{256}$ do not fill the surrounding ambient space uniformly. Instead, their norms concentrate near a relatively thin spherical shell, with empirical mean and standard deviation $\crGeomRbar \pm \crGeomRbarSd$ and 5th--95th percentile range $[\crGeomPFive,\crGeomPNinetyFive]$. This is well below the norm $\sqrt{D-1/2}\approx 16$ of a typical draw from the $\mathcal N(0,I_{256})$ prior. The covariance $\Sigma$ of the encoder means has effective rank (participation ratio) $R_0=(\mathrm{tr}\,\Sigma)^2/\mathrm{tr}(\Sigma^2)=\crGeomParticipation$, and $\crGeomPcNinety$ principal components account for $90\,\%$ of the total variance (Appendix~\ref{app:radius}). We never sample posterior noise during optimization, so we measure the shell radius on encoder means. These statistics describe the shell. They do not guarantee that it is the best place to search for every objective. Relatedly,~\citet{arvanitidis2018latent} show that the latent space of a deep generative model is in general not flat Euclidean and that ignoring this structure can substantially mislead distance-based queries. The empirical concentration of VAE latents on a sphere is consistent with a separate line of work on \emph{hyperspherical} VAEs, which makes this geometry explicit by placing the latent distribution on the unit sphere~\citep{davidson2018hyperspherical,xu2018spherical,decao2020powerspherical}. Our observation can be read as the implicit version of the same phenomenon: rather than redesigning the codec, the \emph{on-sphere scaffold} uses the implicit shell that this Gaussian-prior VAE already exhibits as a constraint on candidate generation. Every newly proposed candidate is placed on this shell, the \emph{empirical sphere},
\begin{equation}
    x = \bar r u, \qquad u \in S^{D-1},
    \label{eq:on_sphere_scaffold}
\end{equation}
where $\bar r$ is the empirical average norm of the encoded cold-start latents within each run and $u$ is a unit direction in the $D$-dimensional latent space. This $\bar r$ is an estimate of the shell radius reported above, and Appendix~\ref{app:radius} compares it with other estimators and with fixed radii. A uniform sample from this scaffold is obtained by drawing $z\sim\mathcal N(0,I_D)$ and returning $\bar r z/\norm{z}$. The scaffold constrains the norm of a candidate; it does not assume that the latent distribution is isotropic. Restricting the directions to the principal subspace of the encoder means did not improve on uniform directions (Appendix~\ref{app:radius}). Random search, \geowalk{}, \geocma{}, and the on-sphere GP variants use the scaffold in Eq.~\eqref{eq:on_sphere_scaffold}, whereas the external baselines use their own candidate generation (Appendix~\ref{app:baselines}).

\paragraph{Acquisition and its maximization.} The GP of the factorial experiment uses the log expected improvement (qLogEI); the acquisition of every other row is given in its label or, where the label has none, in Appendix~\ref{app:baselines}. We compare two ways of maximizing it. With \emph{candidate-set argmax}, each step draws $2{,}048$ fresh uniform candidates from the support and proposes the one with the highest acquisition value. With \emph{continuous maximization}, the acquisition is optimized by L-BFGS-B, the quasi-Newton optimizer of BoTorch~\citep{balandat2020botorch}, from $10$ starts chosen among $512$ raw samples, on both supports. In the box the raw samples are Sobol points of the box, and the optimizer is bounded by the box. On the shell the raw samples are random directions $v\in\R^D$, left unbounded; the acquisition is optimized over $v$ and evaluated at $\bar r v/\norm{v}$, so every iterate lies on the shell. The \emph{box} spans, in each coordinate, the 1st to 99th percentile of $50{,}000$ encoded ZINC-250k molecules. The kernel and its prior are the same on both supports: a Mat\'ern-5/2 kernel with the dimension-scaled lengthscale prior (DSP) of~\citet{hvarfner2024vanilla}. Box candidates are rescaled per coordinate to $[0,1]^D$ before the GP, and shell candidates are not.

\subsection{\geowalk{}: geodesic local search on the empirical sphere}
\label{sec:geowalk}
\geowalk{} is a deliberately minimal optimizer for the \emph{on-sphere scaffold}. It is designed to test how far one can get without fitting a surrogate model once \textbf{candidate generation is restricted to the encoder-mean shell}. Like every optimizer on the molecular benchmark, \geowalk{} starts from the same $N_0=5$ cold-start points. The remaining budget is $T=N-N_0$ evaluations. \geowalk{} first uses $N_1=\lfloor \rho T\rfloor$ of them for uniform exploration of the empirical sphere, where $\rho$ is the exploration ratio. The remaining $T-N_1$ evaluations are used for local geodesic refinement from the best latent point found so far. The two phases balance exploration and exploitation: the uniform samples explore the shell, and the geodesic steps exploit the best point found so far. At the default budget $N=30$, we use $\rho=0.64$, which gives $N_1=16$ exploration samples and $9$ refinement steps.

The refinement phase is a spherical variant of the standard $(1+1)$ evolution strategy. In each refinement step, the algorithm keeps the best point found so far, proposes one new candidate near it on the sphere, and updates the reference point only if the new candidate improves the objective. Let $x_\star$ be the current best latent and let $u_\star=x_\star/\norm{x_\star}$ be its unit direction. To propose a nearby candidate without leaving the sphere, we draw $v\sim\mathcal N(0,I_D)$, remove its radial component, normalize the resulting tangent direction, and take a geodesic step:
\begin{align}
    v &\leftarrow v - \ip{v}{u_\star}u_\star, \qquad v \leftarrow v/\norm v, \label{eq:geowalk_tangent}\\
    u_t &= \cos(\alpha_t)u_\star + \sin(\alpha_t)v, \qquad x_t=\bar r u_t. \label{eq:geowalk_step}
\end{align}
The angle $\alpha_t$ decays linearly from $\alpha_{\max}=0.60$ to $\alpha_{\min}=0.15$ radians during the refinement phase. This creates a simple spherical analogue of a shrinking local search region, but without fitting a GP, learning a covariance matrix, or optimizing an acquisition function.
We also tested an adaptive step-size variant related to the classical $1/5$ success rule~\citep{rechenberg1973,jagerskuepper2007ppsn}. That rule raises the step size when more than about a fifth of recent steps improve the objective, and lowers it otherwise. Unlike that rule, this variant updates the angle after every step: it starts at $0.30$ radians, is multiplied by $1.2$ after an improving step and by $0.85$ otherwise, and stays within $[0.05,0.50]$. The adaptive variant underperforms the fixed decreasing schedule, and so does a constant angle of $0.15$, while constant angles of $0.60$ and $0.35$ show no detected difference from the fixed schedule. We therefore use the fixed angular schedule as a deliberate low-budget design choice rather than as a replacement for full evolution-strategy step-size theory. A full sensitivity analysis of $\rho$ and the angle schedule is reported in Appendix~\ref{app:sensitivity}.

\subsection{\geocma{}: covariance learning in a lifted spherical subspace}
\label{sec:geocma}

\geocma{} is designed to test when the geodesic random walk of \geowalk{} is no longer sufficient. It keeps the same \emph{on-sphere scaffold} from Eq.~\eqref{eq:on_sphere_scaffold}, but replaces both phases of \geowalk{} with covariance learning. The motivation is that \textbf{some search spaces may contain directional structure that can be learned even from a small budget}, so that the sampling distribution on the sphere can be refined toward promising directions.

To do this at low budget, \geocma{} runs Covariance Matrix Adaptation Evolution Strategy (CMA-ES)~\citep{hansen2001cmaes,hansen2019pycma} in a random low-dimensional subspace rather than in the full $D$-dimensional latent space. We draw a Gaussian random matrix and compute an orthonormal basis $Q\in\R^{D\times \tilde{D}}$ by QR decomposition, with $\tilde{D}=16$ in our experiments. CMA-ES works in the projected coordinates $Q^\top x$ of this subspace. There is no uniform exploration phase: its search distribution starts at the projection of the best cold-start point, and the projected cold-start points with their scores are its first update. Each proposal $s\in\R^{\tilde{D}}$ is normalized to the unit sphere in the subspace, lifted to the ambient latent space by $Q$, whose orthonormal columns preserve the norm, and scaled to the empirical radius:
\begin{equation}
    \tilde{s}=\frac{s}{\norm{s}}, \qquad x=\bar r\,Q\tilde{s}.
    \label{eq:geocma_lift}
\end{equation}
On the molecular benchmark, CMA-ES runs with population size $\lambda=5$ and initial step size $\sigma_0=0.5$, and every \geocma{} proposal lies on the same empirical shell as the proposals of \geowalk{}, restricted to the directions of the subspace. After this normalization only the direction of $s$ matters: the covariance learned by CMA-ES shapes which directions are proposed, and its step size sets how widely they spread around the current mean of its search distribution. The two other benchmarks differ in the support. On LassoBench, the CMA-ES sample is lifted without either normalization and clipped to the box $[-1,1]^D$, so there \geocma{} does not search a sphere (Section~\ref{sec:structured}). On NAS-Bench-101, the arch2vec latent space~\citep{yan2020arch2vec} has $16$ dimensions, so the subspace is the whole space and \geocma{} is CMA-ES on the sphere. Full pseudocode for both methods is provided in Appendix~\ref{app:algorithms} (Algorithms~\ref{alg:geowalk} and~\ref{alg:geocma}), and Appendix~\ref{app:sensitivity} compares $\tilde D=8$ and $\tilde D=64$ with the default.

\section{Experiments}
\label{sec:experiments}
The experiments are organized around four questions. (1) How do established optimizers compare with uniform random search on the shell (Section~\ref{sec:rs})? (2) On the shell, does a search without a surrogate differ from a GP (Section~\ref{sec:main_results})? (3) Does the advantage of the shell come from the support, from the acquisition maximizer, or from both, when the two are varied separately (Section~\ref{sec:boxsphere})? (4) When does the random-walk behavior of \geowalk{} stop being sufficient, so that covariance learning in \geocma{} becomes useful (Section~\ref{sec:structured})?

\subsection{Random search on the shell and the external baselines}
\label{sec:rs}
The first experiment tests whether the empirical latent sphere is already a strong search support, even before adding any surrogate model or adaptive local structure. We compare uniform random search on the scaffold in Eq.~\eqref{eq:on_sphere_scaffold} with classical high-dimensional BO and evolutionary baselines, as adapted to this benchmark (Appendix~\ref{app:baselines}), on one matched set (Table~\ref{tab:main}). The comparators are TuRBO~\citep{eriksson2019turbo}, BAxUS~\citep{papenmeier2022baxus} and full-dimensional CMA-ES, the random-embedding methods REMBO and HeSBO, and four published GP recipes (Vanilla BO, Papenmeier MSR, Xu Standard GP and Linear BO), which recent work reports as strong in high dimensions (Section~\ref{sec:related}). We keep the VAE fixed, so methods that update it during the search, such as LOL-BO~\citep{maus2022lolbo} and InvBO~\citep{chu2024invbo}, are not compared. Random search is intentionally minimal: at each step it draws a fresh Gaussian vector, normalizes it to the unit sphere, scales it by $\bar r$, decodes the latent point and evaluates the objective.

\input{tab_main.tex}

The bottom rows of Table~\ref{tab:main} show that this simple on-sphere random search already has a higher mean at $N=30$ than every comparator that proposes candidates off the shell. GaBO and the Riemannian Mat\'ern GP also propose at radius $\bar r$, with kernels defined on the sphere, so they are not counted among the rows off the shell. Random search is ahead on most tasks against each of these comparators. With the task as the unit of the test (Section~\ref{sec:setup}), none of these leads is significant. Table~\ref{tab:main} does not print this test, because its $p$ column tests each row against \geowalk{}, not against random search. The higher mean of random search should not be read as a general criticism of these methods: most were designed for larger budgets or for direct parameter spaces. Appendix~\ref{app:baselines} lists the restrictions under which each method ran.

Against the published recipes, \geowalk{} has the higher mean in every case, but the difference is significant only against one of the two recipes of Papenmeier et al., MLE (scaled) (Table~\ref{tab:published} in Appendix~\ref{app:baselines}). The closest of them is Linear BO~\citep{doumont2026still}, which runs in its authors' box $[-8,8]^D$ from the shared cold-start molecules. \geowalk{} is higher on \crPubLinearVsGeowalkTasksK{} of the 11 tasks, and the difference is not significant. Appendix~\ref{app:convergence} also has our port of Linear BO to the shell.

\subsection{What a surrogate adds on the shell}
\label{sec:main_results}
We next ask whether a Gaussian-process surrogate still provides a substantial advantage once candidate generation is restricted to the same on-sphere scaffold. We compare \geowalk{} with uniform random search on the shell and with the GP of Section~\ref{sec:scaffold} with candidate-set argmax, which we call \emph{the GP on the shell}. It adapts the lengthscale prior of Vanilla BO~\citep{hvarfner2024vanilla} to the shell. It is not a baseline but the scaffold with a surrogate, so Table~\ref{tab:boxsphere} prints it beside its box version, the GP in the box, next to the recipes adapted from Papenmeier MSR and Xu Standard GP. The cross-mean of the GP on the shell is above that of every method in Table~\ref{tab:main}, \geowalk{} included, but its lead over \geowalk{} is not significant, so it is the main comparator of \geowalk{} here.

Earlier in the budget, the GP on the shell is ahead of \geowalk{}, and after thirty evaluations we detect no difference between them. At that budget, \geowalk{} is higher on \crMainVanillaSphereTasksK{} of the 11 tasks. The task-level test does not separate the two ($p=\crMainVanillaSpherePTask{}$), and no per-task difference is significant (Appendix~\ref{app:pertask}). After twenty evaluations of the same runs, the GP on the shell is ahead of \geowalk{} on every task (Appendix~\ref{app:convergence}). At that point \geowalk{} is still in its uniform exploration phase, which ends only after its 21st evaluation (Section~\ref{sec:geowalk}), so until then its runs coincide with random search. The gain of \geowalk{} over uniform random search on the shell at $N=30$ is small but consistent: \geowalk{} is higher than random search on every task (Table~\ref{tab:pertask_full}). The memory and the geodesic refinement steps of \geowalk{} therefore add value beyond the scaffold alone.

\subsection{Candidate support and acquisition maximization}
\label{sec:boxsphere}
Table~\ref{tab:boxsphere} compares three GP recipes on the box and on the shell: a Mat\'ern-5/2 GP with the DSP prior~\citep{hvarfner2024vanilla}, one with the lengthscale initialized at $0.1\sqrt{D}$ as in MSR~\citep{papenmeier2025understanding}, and one initialized at $\sqrt{D}$ as in~\citet{xu2025standardgp}. Each recipe uses L-BFGS-B in the box and candidate-set argmax on the shell, so these rows change the acquisition maximizer together with the support and cannot isolate the support. The $0.1\sqrt{D}$ box row also changes the acquisition function (Appendix~\ref{app:baselines}). A factorial experiment with the DSP-prior GP therefore separates the two factors (Table~\ref{tab:factorial}). It crosses the support, shell or box, with the maximizer: candidate-set argmax or continuous maximization by L-BFGS-B, the same on both supports (Section~\ref{sec:scaffold}). Two of its four cells are the two DSP rows of Table~\ref{tab:boxsphere}. The five contrasts of the factorial, with their task counts and $p$ values, are in Appendix~\ref{app:contrasts}.

\input{tab_boxsphere.tex}
\input{tab_factorial.tex}

The shell helps when the acquisition is maximized over a finite candidate set, and not when it is maximized continuously by L-BFGS-B, which on the shell optimizes over directions and stays on it. Table~\ref{tab:factorial} gives the cross-mean of each cell, and Table~\ref{tab:factorial_contrasts} in Appendix~\ref{app:contrasts} the five contrasts, each with the number of tasks on which it is positive and its $p$ value. With candidate-set argmax, the cross-mean is \crFactorialShellCand{} on the shell and \crFactorialBoxCand{} in the box. The shell is ahead on \crFactorialSupportCandTasksK{} of the 11 tasks, and this difference is significant. With L-BFGS-B, the shell is ahead on \crFactorialSupportContTasksK{} of the 11 tasks, and the difference is not significant. The interaction, the difference between these two support effects, is also significant: on \crFactorialInteractionTasksK{} of the 11 tasks, the lead of the shell over the box is larger under candidate-set argmax than under L-BFGS-B. The maximizer acts in opposite directions on the two supports. On the shell, candidate-set argmax is ahead of L-BFGS-B on most tasks. In the box, L-BFGS-B is ahead of candidate-set argmax on nearly every task. Both of these contrasts are significant.

The recipe rows of Table~\ref{tab:boxsphere} point the same way. For each recipe, the shell version is ahead of the box version on \crBoxVanillaBOTasksK{} of the 11 tasks, but none of these differences is significant.

\subsection{When covariance learning is useful}
\label{sec:structured}
Once candidate generation is restricted to the empirical shell, a simple geodesic random walk shows no detected difference from the GP on the shell after thirty evaluations. On the molecular benchmark, \geocma{} does not improve on \geowalk{}, which is higher on \crCrossGuacamolTasksK{} of the 11 tasks (Table~\ref{tab:crossdomain}). To find where a random walk stops being sufficient, we compare \geowalk{} and \geocma{} on NAS-Bench-101 and LassoBench.

On these two benchmarks we make no claim about a shell, and the sphere is only a fixed-radius search domain. The radius and the clipping of each benchmark are in Appendix~\ref{app:radius}. NAS-Bench-101~\citep{ying2019nasbench} is searched through the 16-dimensional latent space of an arch2vec graph VAE~\citep{yan2020arch2vec} over about $423{,}000$ trained architectures, with CIFAR-10 validation accuracy as the objective. LassoBench~\citep{sehic2022lassobench} is a sparse-regression benchmark with no decoder. We use six of its tasks of increasing dimension: four synthetic tasks, DNA and leukemia. There, \geocma{} clips its points to the box $[-1,1]^D$ without normalizing them, so it does not search a sphere. The column \emph{Best other method} of Table~\ref{tab:crossdomain} is the best of the comparators of Table~\ref{tab:main} and the GP on the shell on GuacaMol, and the best row of Table~\ref{tab:lassobench} on LassoBench. On NAS-Bench-101 it is the best of random search, TuRBO, a GP and full-dimensional CMA-ES.

\input{tab_crossdomain.tex}

Table~\ref{tab:crossdomain} summarizes the boundary. \geowalk{} remains ahead of \geocma{} in mean on the molecular VAE benchmark, where local random movement on the empirical sphere is sufficient. Outside the molecular benchmark, \geocma{} has the better mean of the two, but neither method is detectably ahead of the best other method there. On LassoBench, \geocma{} has a lower normalized loss than \geowalk{} on \crLassoTasksGeocmaK{} of the six tasks, but the task-level test does not detect this difference. The best method there is full-dimensional CMA-ES started at the origin (Appendix~\ref{app:baselines}): it has a lower normalized loss than \geocma{} on five of the six tasks. It also has a lower loss than \geowalk{} on all six tasks, but this sweep is not significant once corrected within the published methods (Table~\ref{tab:lassobench_family}). On NAS-Bench-101, which has a single task and is therefore tested over seeds, \geocma{} improves over \geowalk{}. Random search, however, is ahead of \geowalk{} there, with a Wilcoxon $p$ of \crNasGeowalkVsRandomP{} over the matched seeds, and \geocma{} does not differ detectably from random search.

\section{Discussion}
\label{sec:discussion}

The experiments support a simple interpretation: in the studied molecular VAE benchmark, where the candidates came from mattered more than which surrogate scored them, as long as the acquisition was maximized over a candidate set. Candidates on the encoder-mean shell then outperformed candidates from the box. All three methods on the shell are the on-sphere scaffold with a different search inside it. The GP on the shell is the scaffold with a surrogate, \geowalk{} is the scaffold without one, and \geocma{} is the scaffold with covariance learning. The GP on the shell has the highest cross-mean of any method on this benchmark (Tables~\ref{tab:main} and~\ref{tab:boxsphere}), and \geowalk{} is level with it after thirty evaluations with no surrogate at all. At this budget of thirty evaluations the surrogate is therefore superfluous once the candidates lie on the shell. Earlier in the budget it is ahead, while \geowalk{} is still exploring. This does not make surrogates unnecessary in general: on LassoBench and NAS-Bench-101, \geocma{} has better means than \geowalk{}, while on the molecular benchmark covariance learning does not improve on the walk at $N=30$. Comparisons of latent optimizers should therefore control and report the support and the acquisition maximizer before a gain is attributed to the surrogate.

\paragraph{Limitations.} The advantage of the shell is shown only for candidate sets drawn uniformly from the whole support; local candidate sets, as in trust-region methods, were not tested. The evidence comes from this Gaussian-prior VAE and the 11 GuacaMol tasks of the protocol. In a separate run of the same factorial experiment on nine further GuacaMol tasks, the shell was ahead under candidate-set argmax on only three of them (not shown in the tables). On a second VAE, Chembed~\citep{talibart2026chembed}, whose encoder means also form a shell, moving the GP to the shell gave no detected gain (Appendix~\ref{app:crosscodec}). Locating a shell is therefore no guarantee that searching on it helps.

Budget scaling stops at $N=1000$, in stored runs on ten seeds (Appendix~\ref{app:budget}), so results should not be read asymptotically. The cross-domain evidence for \geocma{} rests mainly on LassoBench, because the NAS-Bench-101 row has only \crNasN{} matched seeds. Other latent spaces were not tested: text VAEs, normalizing flows, diffusion models and the soft-prompt spaces of multimodal language models. Finally, better low-budget molecular optimization could lower experimental cost, but it has potential dual-use risks if applied to harmful objectives.

\paragraph{Conclusion.} In this Gaussian-prior VAE benchmark, shell candidates outperformed box candidates when acquisition was maximized over a candidate set, but not with continuous maximization by L-BFGS-B. The GP on the shell, our scaffold with a surrogate, had the highest cross-mean of any method tested. A surrogate-free geodesic walk on the shell was level with it after thirty evaluations, so at this budget the surrogate is superfluous. Earlier in the budget the surrogate was ahead. This is a controlled study of a Gaussian-prior VAE benchmark, not a recommendation to use VAEs for molecular optimization. The lesson of this study is to control the candidate support and the acquisition maximizer before attributing a gain to the surrogate.

\paragraph{Code and data availability.} The code, run configurations, result files and the generator of every table, figure and number are available at \crRepoUrl. The NAS-Bench-101 embedding is not available; its result files are included. The project page, with the paper, the poster and the results, is \crProjectUrl.

\label{cr:endmain}

\newpage
\begin{ack}
This project started in the Legal Data Analysis Lab, a master's practical course at the Technical
University of Munich; we thank Matthias Grabmair and Mohamed Elganayni for their guidance during the
course. This work was supported by the Czech Science Foundation, grant GA25-15933S, and by the Grant Agency of
the Czech Technical University in Prague, grant SGS24/141/OHK4/3T/14. The experiments were run on the
computing servers of the Institute of Information Theory and Automation, Czech Academy of Sciences.
The authors declare no competing interests.
\end{ack}
\bibliographystyle{unsrtnat}
\bibliography{refs}

\clearpage
\appendix
\raggedbottom %
\makeatletter
\let\cr@apptable\table \let\cr@endapptable\endtable
\renewenvironment{table}[1][H]{\cr@apptable[H]}{\cr@endapptable}
\makeatother
\crtableplacement{htbp}
\section{Detailed pseudocode and settings for \geowalk{} and \geocma{}}
\label{app:algorithms}

Algorithms~\ref{alg:geowalk} and~\ref{alg:geocma} restate the \geowalk{} and \geocma{} procedures described in Sections~\ref{sec:geowalk} and~\ref{sec:geocma} as explicit pseudocode. The cold-start, on-sphere scaffold (Eq.~\ref{eq:on_sphere_scaffold}), and \geocma{} lift step (Eq.~\ref{eq:geocma_lift}) match the main-text description; the inner CMA-ES loop (sample population $\to$ evaluate $\to$ \texttt{tell}) is the standard $(\mu/\mu_w, \lambda)$ generation scheme of~\citet{hansen2001cmaes}, written here in population-per-generation form. The deferred-tell variant used in our implementation evaluates one candidate per step and updates CMA-ES once a full population of size $\lambda$ has accumulated, which is mathematically equivalent. Table~\ref{tab:hyper} lists the settings of \geowalk{} and \geocma{} in the GuacaMol experiments.

\begin{algorithm}[H]\small
\caption{\geowalk{}: geodesic local search on the empirical sphere}
\label{alg:geowalk}
\begin{algorithmic}[1]
\REQUIRE Encoder mean $\mu$, decoder $D_{\mathrm{dec}}$, oracle $f$, latent dimension $D$, budget $N$, cold-start molecules $\{m_i\}_{i=1}^{N_0}$, exploration ratio $\rho$, angles $\alpha_{\max},\alpha_{\min}$
\STATE Encode the cold-start molecules to encoder means and score them directly: $x_i \gets \mu(m_i)$ and $y_i\gets f(m_i)$ for $i=1,\dots,N_0$
\STATE Set $\bar r \gets \tfrac{1}{N_0}\sum_{i=1}^{N_0}\norm{x_i}$ and $T\gets N-N_0$ and $N_1\gets \lfloor \rho T\rfloor$
\FOR{$t=1$ to $N_1$}
    \STATE Draw $z\sim\mathcal N(0,I_D)$ and set $x\gets \bar r z/\norm z$
    \STATE Evaluate $y\gets f(D_{\mathrm{dec}}(x))$ and add $(x,y)$ to the evaluated set
\ENDFOR
\FOR{$t=N_1+1$ to $T$}
    \STATE Let $(x_\star,y_\star)$ be the best evaluated point so far and set $u_\star\gets x_\star/\norm{x_\star}$
    \STATE Draw $v\sim\mathcal N(0,I_D)$ and project it onto the tangent space: $v\gets v-\ip{v}{u_\star}u_\star$
    \STATE Normalize the tangent direction: $v\gets v/\norm v$
    \STATE Set $\alpha_t\gets \alpha_{\max}-(\alpha_{\max}-\alpha_{\min})(t-N_1-1)/\max(T-N_1-1,1)$
    \STATE Propose $x\gets \bar r(\cos\alpha_t\,u_\star+\sin\alpha_t\,v)$
    \STATE Evaluate $y\gets f(D_{\mathrm{dec}}(x))$ and add $(x,y)$ to the evaluated set
\ENDFOR
\RETURN The best evaluated candidate
\end{algorithmic}
\end{algorithm}

\begin{algorithm}[H]\small
\caption{\geocma{}: covariance learning in a lifted spherical subspace}
\label{alg:geocma}
\begin{algorithmic}[1]
\REQUIRE Encoder mean $\mu$, decoder $D_{\mathrm{dec}}$, oracle $f$, latent dimension $D$, budget $N$, cold-start molecules $\{m_i\}_{i=1}^{N_0}$, subspace dimension $\tilde D$, CMA-ES population size $\lambda$, initial step size $\sigma_0$
\STATE Encode the cold-start molecules to encoder means and score them directly: $x_i \gets \mu(m_i)$ and $y_i\gets f(m_i)$ for $i=1,\dots,N_0$
\STATE Set $\bar r \gets \tfrac{1}{N_0}\sum_{i=1}^{N_0}\norm{x_i}$
\STATE Draw a Gaussian random matrix $G\in\R^{D\times\tilde D}$ and form the orthonormal basis $Q\in\R^{D\times\tilde D}$ via the QR decomposition $G=QR$
\STATE Project the cold-start latents into the subspace: $s_i \gets Q^\top x_i$ for $i=1,\dots,N_0$
\STATE Let $s_\star$ be the projected cold-start point with the highest oracle score
\STATE Initialize CMA-ES with mean $s_\star$, step size $\sigma_0$, and population size $\lambda$, then pre-condition it on the cold-start data $\{(s_i, y_i)\}_{i=1}^{N_0}$
\STATE Set $T \gets N - N_0$ and $t \gets 0$
\WHILE{$t < T$}
    \STATE Sample a new population $\{s^{(1)},\dots,s^{(\lambda)}\}$ from CMA-ES and set $K\gets\min(\lambda,\,T-t)$
    \FOR{$k=1$ to $K$}
        \STATE Normalize within the subspace: $\tilde s \gets s^{(k)} / \norm{s^{(k)}}$
        \STATE Lift to the ambient space: $x \gets \bar r\,Q\tilde s$ \quad (the columns of $Q$ are orthonormal, so $\norm{Q\tilde s}=1$)
        \STATE Evaluate $y_k \gets f(D_{\mathrm{dec}}(x))$ and add $(x, y_k)$ to the evaluated set
    \ENDFOR
    \IF{$K=\lambda$}
        \STATE Update CMA-ES with the evaluated population: $\texttt{tell}\bigl(\{(s^{(k)}, y_k)\}_{k=1}^{\lambda}\bigr)$
    \ENDIF
    \STATE $t \gets t + K$
\ENDWHILE
\RETURN The best evaluated candidate
\end{algorithmic}
\end{algorithm}

\input{tab_hyper.tex}

\FloatBarrier
\section{Hyperparameter sensitivity of \geowalk{} and \geocma{}}
\label{app:sensitivity}

\geowalk{} has two design choices. The first is the exploration ratio $\rho$ (default $0.64$, giving $N_1=16$ uniform-sphere samples and $T-N_1=9$ geodesic refinement steps out of the $T=25$ post-cold-start budget). The second is the angle schedule, which decays linearly from $\alpha_{\max}=0.60$ to $\alpha_{\min}=0.15$ radians during the refinement phase. Table~\ref{tab:sensitivity} reports a paired comparison between the default schedule and six variants: three constant-angle variants (covering the start, middle, and end of the default schedule), two exploration-ratio variants ($\rho\approx 0.32$ and $\rho\approx 0.96$), and the success-based rule of Section~\ref{sec:geowalk}. All variants are evaluated on the same $11$ non-degenerate GuacaMol tasks, on one matched set of $\crSensMatchedN$ cells.

Of the three constant-angle variants, only $\alpha=0.15$ (the end of the schedule) underperforms the default linear schedule: it is ahead of the default on only \crSensConstFifteenTasksK{} of the 11 tasks. The constants $\alpha=0.60$ (the beginning of the schedule) and $\alpha=0.35$ (the middle) show no detected difference from the default. The exploration ratio shows an asymmetric pattern: raising $\rho$ from $0.64$ to $\approx 0.96$ (almost all uniform exploration) puts the variant behind the default on all 11 tasks, while reducing $\rho$ to $\approx 0.32$ (less exploration, more refinement) shows no detected difference. The success-based rule is ahead of the default on only \crSensOneFifthTasksK{} of the 11 tasks. Overall, no variant is detectably better than the default, and three are worse.

\paragraph{Subspace dimension of \geocma{}.} The subspace dimension $\tilde D=16$ is a low-budget design choice. With population size $\lambda=5$, the $25$ evaluations give five CMA-ES generations whatever $\tilde D$ is, so $\tilde D$ sets how many directions CMA-ES adapts, not how often it updates. The bottom block of Table~\ref{tab:sensitivity} compares $\tilde D=8$ and $\tilde D=64$ with the default, and neither difference is detected (task-level $p=\crSensDtildeEightPTask$ and $\crSensDtildeSixtyFourPTask$).

\input{tab_sensitivity.tex}

\FloatBarrier
\section{Per-task breakdown of \geowalk{} vs the GP on the shell}
\label{app:pertask}

Section~\ref{sec:main_results} detects no difference between \geowalk{} and the GP on the shell after thirty evaluations. Their paired difference is $\Delta=\crPertaskAllDelta$ at $n=\crPertaskAllN$ matched task-seed cells (last row of Table~\ref{tab:pertask}). Table~\ref{tab:pertask} shows the per-task breakdown across the $11$ non-degenerate GuacaMol tasks, using the same $40$ seeds as Table~\ref{tab:main}.

Per-task absolute differences are small: $|\Delta|\le\crPertaskMaxAbsDelta$ across all $11$ tasks. The largest differences favor the GP on the shell, on Sitagliptin and Osimertinib; the largest in favor of \geowalk{} is on Ranolazine. The signs are balanced across tasks ($\crPertaskTasksGwHigher$ favor \geowalk{}). Only one per-task paired Wilcoxon test reaches $p<0.05$ (Median 1, $p=\crPertaskMedOneP$), and it does not survive Holm correction over the $11$ tasks ($p_{\mathrm{Holm}}=\crPertaskMedOnePHolm$). Cohen's $d$ values are small to moderate in either direction (range $[\crPertaskOsmbD, \crPertaskMedOneD]$). The cross-task mean $\crPertaskAllDelta$ is therefore a representative summary of the per-task picture rather than an artifact of cancellation.

\input{tab_pertask.tex}

\paragraph{Per-task means across the main baselines.} Table~\ref{tab:pertask_full} reports per-task means at $N=30$ across the seven most directly comparable methods (\geowalk{}, \geocma{} and five shell/box baselines). On most tasks, the five methods on the shell (\geowalk{}, \geocma{}, random search on the shell, the GP on the shell, the Linear BO port to the shell) are clustered closely together within each row, while the GP in the box trails (median per-task gap to \geowalk{} is \crFullMedianBoxGap{}). On Median Molecules~1 and Sitagliptin, however, the two box columns have the highest means.

\input{tab_pertask_full.tex}

\FloatBarrier
\section{Per-task convergence on the GuacaMol benchmark}
\label{app:convergence}

\begin{figure}[p]
\centering
\includegraphics[width=0.985\textwidth]{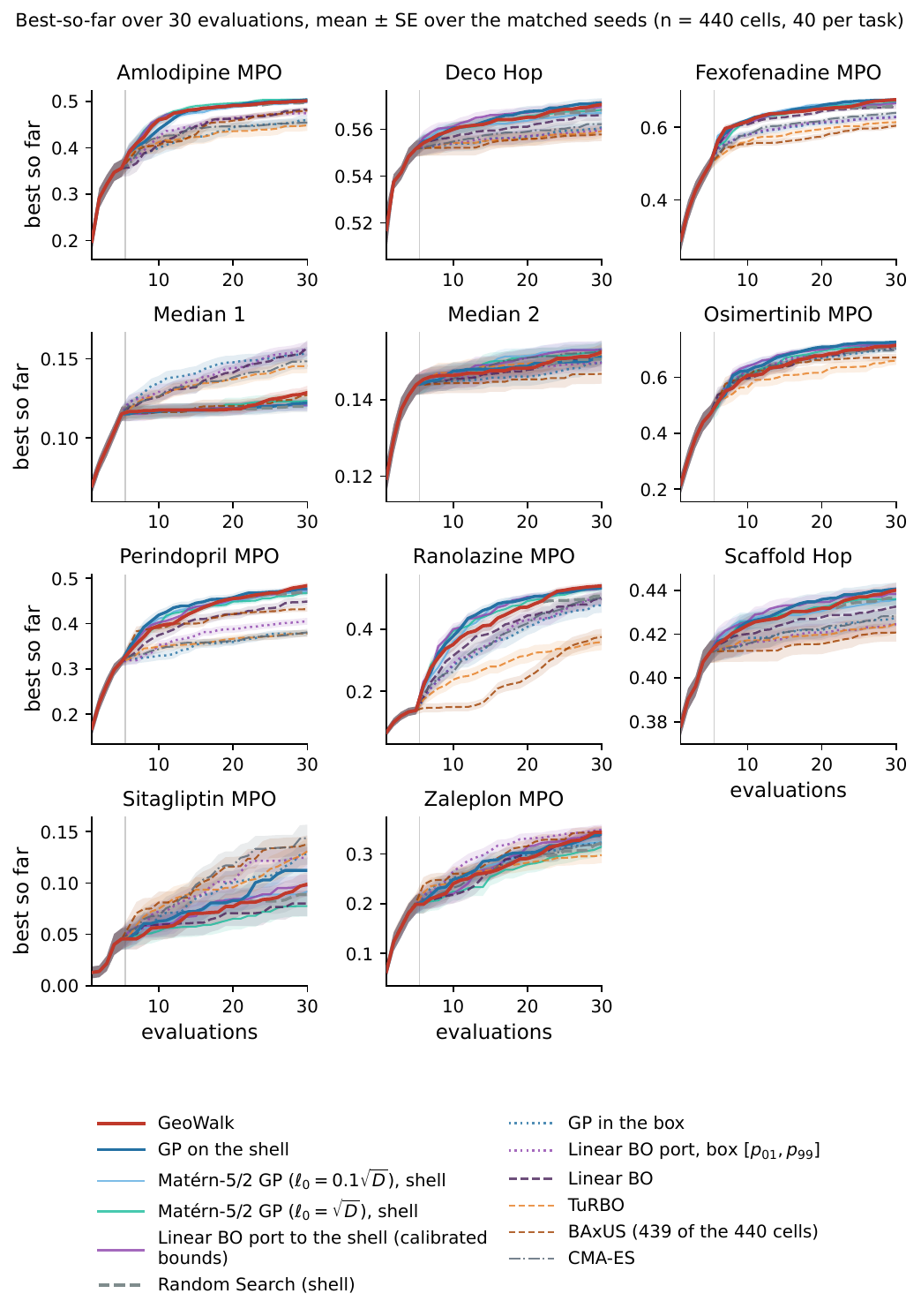}
\caption{\pb{Thirty-evaluation convergence on the 11 non-degenerate GuacaMol tasks.} Curves show the best score found so far over $30$ evaluations, averaged over the seeds of one matched set of $\crFigConvergenceN$ cells; bands indicate standard error. The grey vertical line marks the end of the five shared cold-start molecules. The three curves drawn heavier are \geowalk{}, the GP on the shell and random search on the shell, which Section~\ref{sec:main_results} compares. The legend uses the names of Section~\ref{sec:main_results} and Tables~\ref{tab:main} and~\ref{tab:boxsphere}. The two ports of Linear BO appear in the appendix only (Tables~\ref{tab:pertask_full}, \ref{tab:trajectories}, \ref{tab:lassobench} and~\ref{tab:baselines}).}
\label{fig:convergence}
\end{figure}

Figure~\ref{fig:convergence} shows the full $30$-evaluation convergence traces for each of the $11$ non-degenerate GuacaMol tasks, complementing the aggregate cross-task numbers in Table~\ref{tab:main}. Table~\ref{tab:trajectories} aggregates the same runs over tasks. It names its rows as Table~\ref{tab:main} and Figure~\ref{fig:convergence} do. The Linear BO port to the shell, which Table~\ref{tab:main} does not print (Section~\ref{sec:rs}), changes the support, the acquisition (qLogEI instead of LogEI) and its maximization, so we do not read the difference between the port and the published recipe as an effect of the support alone. The kernel bounds of the port were calibrated after a first run (Appendix~\ref{app:baselines}). The first 16 proposals of \geowalk{} are uniform draws under the same seed as random search, so the two rows of Table~\ref{tab:trajectories} agree at $N=10$ and $N=20$. The GP on the shell is ahead of \geowalk{} at $N=20$ on all 11 tasks, and \geowalk{} catches up only in its last evaluations (Table~\ref{tab:trajectories}, bottom). In the aggregate, the GP rows on the shell are above the GP rows in the box at each budget of Table~\ref{tab:trajectories}. These differences were not tested.

\input{tab_trajectories.tex}

\FloatBarrier
\section{Scope of the codec comparison}
\label{app:crosscodec}

The main experiments use the SELFIES Transformer VAE released with LOL-BO~\citep{maus2022lolbo}. Stored runs exist for two further molecular codecs: Chembed~\citep{talibart2026chembed}, which is a second pre-trained Transformer VAE, and a Power-Spherical VAE~\citep{decao2020powerspherical}. For each codec we evaluated four methods at the same $N=30$ budget. Table~\ref{tab:crosscodec_means} reports the cross-task means.

\input{tab_crosscodec.tex}

The comparison does not extend the main result beyond the SELFIES VAE. The Power-Spherical VAE normalizes its latent inside the decoder and decodes stochastically, so it cannot test where candidates are placed radially. Chembed's encoder means also form a shell (the norm statistics on ZINC-250k are in the results bundle). Even so, on Chembed the GP on the shell is not detectably ahead of the GP in the box: it is ahead on \crCodecChembedSphereBoxTasksK{} of the 11 tasks (task-level $p=\crCodecChembedSphereBoxPTask$). On Chembed the GP on the shell also has a higher mean than \geowalk{}. \geowalk{} is higher on only \crCodecChembedGwVsSphereTasksK{} of the 11 tasks, and the difference is not significant. A systematic study across more diverse codecs (e.g., diffusion-based or normalizing-flow latent spaces) is left to future work.

\FloatBarrier
\section{Budget scaling beyond \texorpdfstring{$N=30$}{N=30}}
\label{app:budget}

The main experiments focus on the low-budget regime $N=30$ motivated by wet-lab cost structures, but it is useful to ask whether the same pattern persists when more evaluations are available. Table~\ref{tab:budget} reports stored runs with $N=1000$, read at budgets $N\in\{30,50,100,200,500,1000\}$ on the same $11$ non-degenerate GuacaMol tasks, using seeds $42$--$51$. The table is descriptive; the goal is not to claim asymptotic dominance. \geowalk{} and the GP on the shell remain close across all budgets, with matched differences of at most \crBudgetVsVanillaSphereNOneZeroZeroZeroDelta{} for $N\ge 50$ and \crBudgetVsVanillaSphereNThreeZeroDelta{} at $N=30$. \geocma{} and random search on the shell also improve steadily with budget, but remain below \geowalk{} and, except for \geocma{} at $N=100$ and $N=200$, below the GP on the shell. %
The box row, a Thompson-sampling GP in the bounding box of the cold-start latents, is below every method on the shell at every budget. It differs from them in more than the support, so these runs carry no claim about the support.

\input{tab_budget.tex}

\FloatBarrier
\section{LassoBench per dimension}
\label{app:lassobench_phase}

Table~\ref{tab:lassobench} accompanies the LassoBench discussion in Section~\ref{sec:structured} by showing the per-dimension performance behind the aggregate row in Table~\ref{tab:crossdomain}. The GP rows other than TuRBO and BAxUS maximize the acquisition continuously with L-BFGS-B over the box $[-1,1]^D$. TuRBO and BAxUS draw a Thompson sample over a finite candidate set inside a trust region. For BAxUS, the trust region lies in its embedded subspace. \geocma{} has a lower normalized loss than \geowalk{} at \crLassoTasksGeocmaK{} of the six dimensions. It has the lowest loss of all methods only at $D=7129$. At the other dimensions, the CMA-ES recipe of LassoBench (all $D$ dimensions, started at the origin) has a lower mean loss than \geocma{}. No difference between \geowalk{} and a published method is significant after correction within the published methods (Table~\ref{tab:lassobench_family}). The runs also reveal a practical limitation of full-dimensional GPs at very low budget. Several GP methods lack seeds at some dimensions, because a model fit failed or, for MSR and MLE (scaled) at $D=1000$, because runs exceeded the limit of one hour. Both \geowalk{} and \geocma{} avoid the full-dimensional GP fitting problem altogether.

\input{tab_lassobench.tex}

\FloatBarrier

\section{Baseline implementation details}
\label{app:baselines}

This appendix summarizes every method of Tables~\ref{tab:main} and~\ref{tab:lassobench} as it ran. %
Table~\ref{tab:baselines} gives one row per method: its implementation, surrogate, acquisition and maximizer, and candidate support.
The box version of the $0.1\sqrt{D}$ recipe of Table~\ref{tab:boxsphere} uses the acquisition of MSR, analytic LogEI with sample-around-best restarts (RAASP in Table~\ref{tab:baselines}), so that row differs from its shell version in the acquisition as well as in the support and the maximizer. We print that row because it is the published call. A box run of this recipe with qLogEI, which has a higher mean than the printed box row, is in the full table in the repository. Neither row enters the factorial of Table~\ref{tab:factorial}, which uses the DSP recipe with the same maximizer on both supports.

The full card of every method is in the repository (\texttt{docs/baselines/BASELINE\_CARDS.md}). Each card has ten fields: the software version, kernel, acquisition and maximizer, lengthscale initialization, lengthscale prior, candidate support, trust region or RAASP settings, seeds, repository of the published method, and every deviation from it. 
\input{tab_main_notes.tex}

\input{tab_baselines_overview.tex}

\input{tab_published.tex}

\FloatBarrier

\section{Shell radius and alternative candidate supports}
\label{app:radius}

\begin{figure}[H]
\centering
\includegraphics{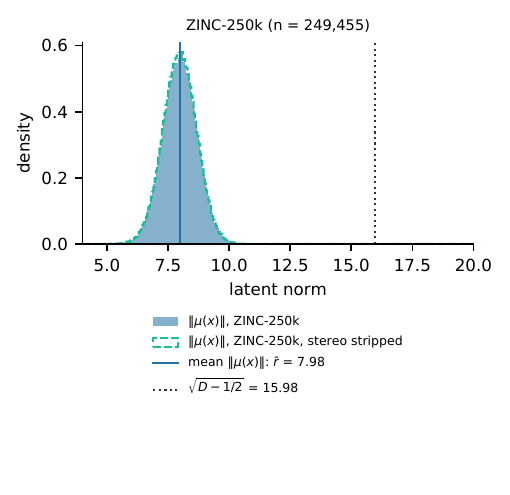}
\caption{\pb{Norms of the encoder means on ZINC-250k.} Histogram of $\|\mu(x)\|$ for the $\crFigShellsN$ molecules, encoded with the default pipeline (out-of-vocabulary stereo tokens dropped) and with stereochemistry stripped before encoding. The marks show the mean norm over ZINC-250k and $\sqrt{D-1/2}\approx16$, the typical norm of a sample from the $\mathcal N(0,I_D)$ prior with $D=256$.}
\label{fig:shells}
\end{figure}

Table~\ref{tab:geometry} and Figure~\ref{fig:shells} describe the shell of the encoder means on ZINC-250k. Let $\Sigma$ and $\bar\mu$ be the covariance and the mean of the encoder means, and $m=\mathrm{tr}\,\Sigma+\|\bar\mu\|^2$. The second-order approximation of the mean norm, $\sqrt m-\mathrm{Var}(\|\mu\|^2)/(8m^{3/2})$, differs from the measured mean norm by $\crGeoTabRadiusModelRelErrPctFull\,\%$. It is computed on the same encodings, so it describes the shell rather than predicting it. By default, stereo tokens outside the vocabulary of the VAE are dropped before encoding, and the affected atoms are lost. Every run of the paper uses this default pipeline (Section~\ref{sec:setup}). Stripping stereochemistry with RDKit before encoding keeps these atoms, and the mean norm then changes by $\crGeomStripShiftPct\,\%$.

\input{tab_geometry.tex}

\paragraph{Radius, estimator and thickness.} The three blocks of Table~\ref{tab:radius} vary how \geowalk{} sets the radius of its proposals. Fixed radii from $6$ to $12$ show no detected difference from $r=8$ at the task level. A radius of $4$ is ahead on only \crRadiusRFourTasksK{} of the 11 tasks, and a radius of $16$ on only \crRadiusROneSixTasksK{}. Both have lower means, but neither difference is significant at the task level. Setting the radius to the median of the five cold-start norms, instead of their mean, gives a lower mean score. This variant is ahead on only \crEstMedianTasksK{} of the 11 tasks (task-level $p=\crEstMedianPTask$). A fixed radius equal to the mean norm over ZINC-250k and the three thick shells show no detected difference from the default.

\input{tab_radius.tex}

\paragraph{Supports inside the latent cloud and directions.} Every convex sampler and the Gaussian fit of Table~\ref{tab:convex} has a lower mean than uniform sampling on the shell (the convex samplers draw Dirichlet combinations of $k$ random latents, not uniform samples of the convex hull), with differences from \crConvexKThreeDelta{} to \crConvexKSixFourDelta. With the task as the unit, none of these differences is significant. Averaging, which moves candidates inside the shell, does not explain this gap: the sampler with $k=1$ draws encoded molecules themselves, whose mean norm equals the shell radius, and it is about as weak as the sampler with $k=64$. Two controls change the directions at the shell radius (Table~\ref{tab:convex}). Random search restricted to the top-64 principal subspace has a lower mean than random search on the full shell, ahead on only \crDirPcaTasksK{} of the 11 tasks (task-level $p=\crDirPcaPTask$). The GP with candidate-set argmax over directions drawn from a cube centered at the origin and rescaled to $\bar r$ shows no detected difference from the GP on the shell (task-level $p=\crDirAngularPTask$). These directions come from that cube, not from the per-coordinate box of Section~\ref{sec:boxsphere}. These controls suggest that, among the tested alternatives, matching the radial scale is more important than explicitly imposing the observed directional structure.

\input{tab_convex.tex}

\paragraph{Radius and clipping on NAS-Bench-101 and LassoBench.} On NAS-Bench-101, both methods search a sphere whose radius is the mean norm of the embeddings, and \geocma{} starts with a step size equal to the standard deviation of its projected cold-start points. On LassoBench, \geowalk{} uses the mean norm of its cold-start points as the radius and clips its proposals to the box $[-1,1]^D$. \geocma{} clips its lifted points to the same box without normalizing them, so there it does not search a sphere.

\FloatBarrier

\section{The five contrasts of the factorial experiment}
\label{app:contrasts}
Table~\ref{tab:factorial_contrasts} lists the five contrasts of the factorial experiment of Section~\ref{sec:boxsphere}: the support effect under each maximizer, the maximizer effect on each support, and their interaction, the difference between the two support effects. Each contrast is the difference of two cross-means of Table~\ref{tab:factorial}, tested at the task level and corrected within the five contrasts (Section~\ref{sec:setup}). Section~\ref{sec:boxsphere} quotes the task counts of the factorial from this table.

\input{tab_factorial_contrasts.tex}

\section{Compute and assets}
\label{app:compute}

\paragraph{Machines.} Servers~A and~B each have two NVIDIA L40S GPUs (48\,GB each) and two Intel Xeon Gold 6542Y CPUs. Server~A ran the GuacaMol runs other than the stored runs and the two sphere-manifold GPs. Server~B ran these two GPs, all LassoBench runs and the checks of the published recipes. All of these runs used only the CPUs of server~B. The stored runs were recorded on server~B and on server~C. Server~C has two Intel Xeon Gold 5315Y CPUs, an NVIDIA A100 GPU (80\,GB) and an NVIDIA RTX A5000 GPU.

\paragraph{Compute.} A table in the repository lists, for each block of runs, its machine, its process-hours and the median time of one run. A process-hour is one hour of wall-clock time as recorded in a result file, and the totals sum these hours over the runs of a machine. The runs other than the stored runs took about \crComputeGpuServerHours{} process-hours on server~A and about \crComputeCpuServerHours{} on server~B. Of the hours on server~B, about \crComputeVerifyHours{} were runs that checked the published recipes; the outcome of each check is in the results bundle, with the method cards. The stored runs that the printed tables read took about \crComputeStoredPrintedHours{} process-hours. This total counts only the runs whose result files record a time. The project also used compute for preliminary and failed runs that the paper does not report.

\paragraph{Assets and licenses.} Every external asset is cited where it is used. The list below gives the license of each asset and its version where it has one. The third-party notices of the repository give the commit of every asset and list every third-party file with its license.
\begin{itemize}
\item \emph{Molecules and models.} The SELFIES~\citep{krenn2020selfies} VAE and code released with LOL-BO~\citep{maus2022lolbo,maus2022lolbocode} are MIT. GuacaMol~\citep{brown2019guacamol} 0.5.2 is MIT; its data are CC BY 4.0, derived from ChEMBL under CC BY-SA 3.0. ZINC-250k~\citep{irwin2012zinc}, in the subset of~\citet{gomez2018automatic}, is free to use without a formal license; the file comes from the Apache-2.0 \texttt{chemical\_vae} repository. The code of Chembed~\citep{talibart2026chembed} is MIT. The weights of Chembed declare no license, and our repository does not redistribute them. We trained the Power-Spherical VAE of Appendix~\ref{app:crosscodec} with the \texttt{power\_spherical} package of~\citet{decao2020powerspherical}, which is MIT.
\item \emph{Benchmarks.} LassoBench~\citep{sehic2022lassobench} is MIT for the code and the synthetic tasks; its real-world DNA and leukemia data fall under the terms of the LIBSVM website. NAS-Bench-101~\citep{ying2019nasbench} and arch2vec~\citep{yan2020arch2vec} are Apache-2.0.
\item \emph{Software.} PyTorch~\citep{paszke2019pytorch} 2.11.0, pycma~\citep{hansen2019pycma} 4.4.4 and RDKit~\citep{rdkitsoftware} 2026.03.1 are BSD-3. BoTorch~\citep{balandat2020botorch} 0.17.2 and GPyTorch~\citep{gardner2018gpytorch} 1.15.2 are MIT, and selfies 2.2.0 is Apache-2.0. The released code of Linear BO~\citep{doumont2026still}, GaBO~\citep{jaquier2020gabo} and the Riemannian Mat\'ern GP~\citep{jaquier2022geometry} is MIT.
\item \emph{Ported code.} Our port of the official TuRBO code~\citep{eriksson2019turbo} (\url{https://github.com/uber-research/TuRBO}) is used for every TuRBO row of the paper and for the TuRBO curve of Figure~\ref{fig:convergence}. That code is under the TuRBO license, which is non-commercial, so the port is distributed under the same license, with the change notice it requires. The BAxUS row of Table~\ref{tab:lassobench} ports the BoTorch BAxUS tutorial, which is MIT; this port shares its file with the TuRBO port and is distributed with it under the TuRBO license.
\item \emph{Our code} is released under the MIT license, except the file with the two ports above.
\end{itemize}

\clearpage
\input{checklist.tex}

\end{document}

%% file: cr_status.tex
\crprelimfalse

%% file: numbers.tex
\newcommand{\crFigConvergenceN}{\ensuremath{440}}

\newcommand{\crFigShellsN}{\ensuremath{249{,}455}}

\newcommand{\crGeomNZinc}{\ensuremath{249{,}455}}
\newcommand{\crGeomPFive}{\ensuremath{6.81}}
\newcommand{\crGeomPNinetyFive}{\ensuremath{9.14}}
\newcommand{\crGeomParticipation}{\ensuremath{64.7}}
\newcommand{\crGeomPcNinety}{\ensuremath{66}}

\newcommand{\crGeomRbar}{\ensuremath{7.98}}

\newcommand{\crGeomRbarSd}{\ensuremath{0.71}}

\newcommand{\crGeomStripShiftPct}{\ensuremath{-0.10}}

\newcommand{\crGeomTokenLossPct}{\ensuremath{63.1}}

\newcommand{\crBaselinesGaboCells}{\ensuremath{440}}

\newcommand{\crBaselinesGaboFallbackCells}{\ensuremath{254}}
\newcommand{\crBaselinesGaboFallbackSteps}{\ensuremath{4{,}420}}

\newcommand{\crBoxVanillaBOTasksK}{\ensuremath{9}}

\newcommand{\crFactorialBoxCand}{\ensuremath{0.3925}}

\newcommand{\crFactorialInteractionTasksK}{\ensuremath{9}}

\newcommand{\crFactorialShellCand}{\ensuremath{0.4228}}

\newcommand{\crFactorialSupportCandTasksK}{\ensuremath{9}}

\newcommand{\crFactorialSupportContTasksK}{\ensuremath{6}}

\newcommand{\crBudgetVsVanillaSphereNOneZeroZeroZeroDelta}{\ensuremath{+0.008}}
\newcommand{\crBudgetVsVanillaSphereNThreeZeroDelta}{\ensuremath{-0.002}}

\newcommand{\crComputeCpuServerHours}{\ensuremath{4{,}372}}
\newcommand{\crComputeGpuServerHours}{\ensuremath{647}}

\newcommand{\crComputeStoredPrintedHours}{\ensuremath{1{,}289}}
\newcommand{\crComputeVerifyHours}{\ensuremath{4{,}185}}

\newcommand{\crConvexKSixFourDelta}{\ensuremath{-0.0295}}

\newcommand{\crConvexKThreeDelta}{\ensuremath{-0.0209}}

\newcommand{\crCodecChembedGwVsSphereTasksK}{\ensuremath{4}}

\newcommand{\crCodecChembedSphereBoxPTask}{\ensuremath{0.52}}

\newcommand{\crCodecChembedSphereBoxTasksK}{\ensuremath{6}}

\newcommand{\crCrossGuacamolTasksK}{\ensuremath{9}}

\newcommand{\crNasGeowalkVsRandomP}{\ensuremath{0.037}}
\newcommand{\crNasN}{\ensuremath{10}}

\newcommand{\crEstMedianPTask}{\ensuremath{0.083}}

\newcommand{\crEstMedianTasksK}{\ensuremath{2}}

\newcommand{\crGeoTabRadiusModelRelErrPctFull}{\ensuremath{+0.03}}

\newcommand{\crLassoTasksGeocmaK}{\ensuremath{5}}

\newcommand{\crMainVanillaSpherePTask}{\ensuremath{0.83}}

\newcommand{\crMainVanillaSphereTasksK}{\ensuremath{6}}

\newcommand{\crDirAngularPTask}{\ensuremath{0.90}}

\newcommand{\crDirPcaPTask}{\ensuremath{0.17}}

\newcommand{\crDirPcaTasksK}{\ensuremath{3}}

\newcommand{\crPertaskAllDelta}{\ensuremath{-0.0001}}

\newcommand{\crPertaskAllN}{\ensuremath{440}}

\newcommand{\crPertaskMaxAbsDelta}{\ensuremath{0.0135}}
\newcommand{\crPertaskMedOneD}{\ensuremath{+0.40}}

\newcommand{\crPertaskMedOneP}{\ensuremath{0.024}}
\newcommand{\crPertaskMedOnePHolm}{\ensuremath{0.27}}

\newcommand{\crPertaskOsmbD}{\ensuremath{-0.21}}

\newcommand{\crPertaskTasksGwHigher}{\ensuremath{6/11}}

\newcommand{\crFullMedianBoxGap}{\ensuremath{-0.013}}

\newcommand{\crPubLinearVsGeowalkTasksK}{\ensuremath{10}}

\newcommand{\crRadiusRFourTasksK}{\ensuremath{3}}

\newcommand{\crRadiusROneSixTasksK}{\ensuremath{3}}

\newcommand{\crSensConstFifteenTasksK}{\ensuremath{2}}

\newcommand{\crSensDtildeEightPTask}{\ensuremath{0.21}}

\newcommand{\crSensDtildeSixtyFourPTask}{\ensuremath{0.70}}

\newcommand{\crSensMatchedN}{\ensuremath{440}}

\newcommand{\crSensOneFifthTasksK}{\ensuremath{1}}

%% file: tab_main.tex
\ifdefined\crtabbox\else\newsavebox{\crtabbox}\fi
\begin{table}[!t]
\centering
\caption{\textbf{Cross-task means and paired statistics for \geowalk{} versus all comparators at $N=30$} on one matched set of $n=440$ (task, seed) cells of the 11 GuacaMol tasks. REMBO and HeSBO propose on the empirical shell; GaBO and Riemannian Mat\'ern GP propose on the sphere manifold; all other comparators propose off the shell. $\Delta$ is the improvement of \geowalk{} over the comparator. W/L counts the matched cells where \geowalk{} wins or loses, excluding ties. $p_{\mathrm{task}}$ is the exact two-sided Wilcoxon test on the 11 per-task mean differences, not corrected. $d$ is Cohen's $d$ of the paired cell differences. Bold marks the highest mean. Notes are in Appendix~\ref{app:baselines}.}
\label{tab:main}
\small
\begin{lrbox}{\crtabbox}
\begin{tabularx}{\linewidth}{>{\raggedright\arraybackslash}Xrrrrr}
\toprule
Method & Cross-mean & $\Delta$ & W/L & $p_{\mathrm{task}}$ & $d$ \\
\midrule
CMA-ES~\citep{hansen2001cmaes,hansen2019pycma} & $0.4026$ & $+0.0200$ & 253/148 & $0.12$ & $0.25$ \\
TuRBO~\citep{eriksson2019turbo} & $0.3790$ & $+0.0437$ & 270/123 & $0.032$ & $0.43$ \\
BAxUS$^{\ddagger}$~\citep{papenmeier2022baxus} & $0.3907$ & $+0.0320$ & 283/101 & $0.024$ & $0.34$ \\
REMBO~\citep{wang2016rembo} & $0.4092$ & $+0.0135$ & 238/144 & $0.002$ & $0.23$ \\
HeSBO~\citep{nayebi2019hesbo} & $0.4120$ & $+0.0106$ & 225/158 & $0.032$ & $0.17$ \\
Vanilla BO~\citep{hvarfner2024vanilla} & $0.4008$ & $+0.0218$ & 238/159 & $0.083$ & $0.26$ \\
Papenmeier MSR~\citep{papenmeier2025understanding} & $0.3899$ & $+0.0327$ & 281/109 & $0.019$ & $0.40$ \\
Xu Standard GP~\citep{xu2025standardgp} & $0.3951$ & $+0.0275$ & 270/120 & $0.042$ & $0.37$ \\
Linear BO~\citep{doumont2026still} & $0.4101$ & $+0.0125$ & 226/172 & $0.032$ & $0.17$ \\
GaBO~\citep{jaquier2020gabo} & $0.4153$ & $+0.0074$ & 231/165 & $0.17$ & $0.13$ \\
Riemannian Mat\'ern GP~\citep{jaquier2022geometry} & $0.4059$ & $+0.0167$ & 230/152 & $0.014$ & $0.26$ \\
\midrule
\multicolumn{6}{@{}l}{\textit{This paper, on the empirical shell (Eq.~\ref{eq:on_sphere_scaffold}):}} \\
Random Search (shell) & $0.4118$ & $+0.0108$ & 194/64 & $0.00098$ & $0.30$ \\
\geocma{} & $0.4158$ & $+0.0068$ & 225/169 & $0.054$ & $0.11$ \\
\geowalk{} & $\mathbf{0.4226}$ & -- & -- & -- & -- \\
\bottomrule
\end{tabularx}
\end{lrbox}
\ifdim\wd\crtabbox>\linewidth\resizebox{\linewidth}{!}{\usebox{\crtabbox}}\else\usebox{\crtabbox}\fi
\end{table}

%% file: tab_boxsphere.tex
\ifdefined\crtabbox\else\newsavebox{\crtabbox}\fi
\begin{table}[t]
\centering
\caption{\textbf{Box versus empirical shell within each GP recipe.} Each GP on the empirical shell (candidate-set argmax) and in the box $[p_{01},p_{99}]$ (L-BFGS-B), on one matched set of $n=440$ cells; cross-means as in Table~\ref{tab:main}. Tasks counts the tasks, of the 11, on which the shell version has the higher mean. $p_{\mathrm{task}}^{\mathrm{BH}}$ is the task-level test of the shell version against the box version, corrected within the three recipes. The two DSP rows are the GP on the shell and the GP in the box of Sections~\ref{sec:main_results} and~\ref{sec:boxsphere}.}
\label{tab:boxsphere}
\small
\begin{lrbox}{\crtabbox}
\begin{tabularx}{\linewidth}{>{\raggedright\arraybackslash}Xlrrr}
\toprule
Variant & Candidate support & Cross-mean & \begin{tabular}[b]{@{}c@{}}Tasks\\shell\\higher\end{tabular} & $p_{\mathrm{task}}^{\mathrm{BH}}$ \\
\midrule
Mat\'ern-5/2 GP (DSP prior), shell & Empirical shell & $0.4228$ & 9/11 & $0.081$ \\
Mat\'ern-5/2 GP (DSP prior), box & $[p_{01},p_{99}]$ & $0.3989$ &  &  \\
\midrule
Mat\'ern-5/2 GP ($\ell_0=0.1\sqrt{D}$), shell & Empirical shell & $0.4176$ & 9/11 & $0.081$ \\
Mat\'ern-5/2 GP ($\ell_0=0.1\sqrt{D}$), box & $[p_{01},p_{99}]$ & $0.3845$ &  &  \\
\midrule
Mat\'ern-5/2 GP ($\ell_0=\sqrt{D}$), shell & Empirical shell & $0.4149$ & 9/11 & $0.083$ \\
Mat\'ern-5/2 GP ($\ell_0=\sqrt{D}$), box & $[p_{01},p_{99}]$ & $0.3957$ &  &  \\
\bottomrule
\end{tabularx}
\end{lrbox}
\ifdim\wd\crtabbox>\linewidth\resizebox{\linewidth}{!}{\usebox{\crtabbox}}\else\usebox{\crtabbox}\fi
\end{table}

%% file: tab_factorial.tex
\ifdefined\crtabbox\else\newsavebox{\crtabbox}\fi
\begin{table}[t]
\centering
\caption{\textbf{The support $\times$ acquisition-maximization factorial: cross-mean of each cell.} One Mat\'ern-5/2 GP (DSP prior) on $n=440$ matched cells; continuous maximization is L-BFGS-B on both supports. Cross-means as in Table~\ref{tab:main}. The five contrasts of these cells, with their task counts and corrected $p$ values, are in Table~\ref{tab:factorial_contrasts} (Appendix~\ref{app:contrasts}).}
\label{tab:factorial}
\small
\begin{lrbox}{\crtabbox}
\begin{tabular}{lrr}
\toprule
Support & Candidate-set argmax & Continuous (L-BFGS-B) \\
\midrule
Empirical shell & $0.4228$ & $0.3999$ \\
Box $[p_{01},p_{99}]$ & $0.3925$ & $0.3989$ \\
\bottomrule
\end{tabular}
\end{lrbox}
\ifdim\wd\crtabbox>\linewidth\resizebox{\linewidth}{!}{\usebox{\crtabbox}}\else\usebox{\crtabbox}\fi
\end{table}

%% file: tab_crossdomain.tex
\ifdefined\crtabbox\else\newsavebox{\crtabbox}\fi
\begin{table}[t]
\centering
\caption{\textbf{\geowalk{} and \geocma{} across benchmark regimes.} $N=30$. Arrows give the optimization direction, and bold marks the better of the two means. GuacaMol: mean of per-task means on the matched cells. LassoBench: each method's loss divided by the random-search mean loss of the task, averaged over the six tasks. $n$ is the number of paired cells, or of paired seeds on NAS-Bench-101. Best other method: the best mean among the other methods that cover the row's matched set. $\Delta$ is paired and favors the better method, $|d|$ is Cohen's $d$ of the paired cells, and $p_{\mathrm{task}}$ is the task-level test of Section~\ref{sec:setup}, uncorrected across the rows; $^a$ marks the paired test over seeds, because NAS-Bench-101 has a single task.}
\label{tab:crossdomain}
\small
\setlength{\tabcolsep}{3pt}
\begin{lrbox}{\crtabbox}
\begin{tabularx}{\linewidth}{>{\raggedright\arraybackslash}Xcrrrrrr>{\raggedright\arraybackslash}p{2.2cm}}
\toprule
Benchmark & Dir. & \geowalk{} & \geocma{} & $\Delta$ & $|d|$ & $p_{\mathrm{task}}$ & $n$ & Best other method \\
\midrule
GuacaMol mol VAE (256D) & $\uparrow$ & $\mathbf{0.423}$ & $0.416$ & $0.007$ & $0.11$ & $0.054$ & 440 & GP on the shell, $0.423$ \\
NAS-Bench-101 graph VAE (16D) & $\uparrow$ & $0.937$ & $\mathbf{0.942}$ & $0.004$ & $0.94$ & $0.020^{a}$ & 10 & random search, $0.941$ \\
LassoBench, $D\in[60,7129]$, normalized & $\downarrow$ & $0.796$ & $\mathbf{0.495}$ & $0.300$ & $0.37$ & $0.094$ & 240 & CMA-ES, $0.428$ \\
\bottomrule
\end{tabularx}
\end{lrbox}
\ifdim\wd\crtabbox>\linewidth\resizebox{\linewidth}{!}{\usebox{\crtabbox}}\else\usebox{\crtabbox}\fi
\end{table}

%% file: tab_hyper.tex
\ifdefined\crtabbox\else\newsavebox{\crtabbox}\fi
\begin{table}[t]
\centering
\caption{\textbf{Settings of \geowalk{} and \geocma{} on GuacaMol.} These settings hold for every task and seed. The settings that differ on LassoBench and NAS-Bench-101 are given in Section~\ref{sec:structured} and Appendix~\ref{app:radius}.}
\label{tab:hyper}
\footnotesize
\setlength{\tabcolsep}{4pt}
\begin{lrbox}{\crtabbox}
\begin{tabularx}{\linewidth}{>{\raggedright\arraybackslash}p{4.0cm}>{\raggedright\arraybackslash}X}
\toprule
Setting & Value \\
\midrule
\multicolumn{2}{@{}l}{\textit{\textup{\geowalk{}}}} \\
Cold-start points $N_0$ & 5 \\
Evaluations after the cold start $T$ & 25 \\
Budget $N=N_0+T$ & 30 \\
Exploration ratio $\rho$ & 0.64 \\
Exploration steps $N_1=\lfloor\rho T\rfloor$ & 16 uniform proposals on the shell \\
Refinement steps $T-N_1$ & 9 geodesic steps \\
Angle $\alpha_{\max}\to\alpha_{\min}$ & $0.60\to0.15$ rad \\
Angle schedule & linear in the refinement step, over $\max(T-N_1-1,1)=8$ intervals \\
Anchor of a refinement step & best evaluated point so far, cold starts included \\
Radius $\bar r$ & mean norm of the $N_0$ cold-start latents \\
\multicolumn{2}{@{}l}{\textit{\textup{\geocma{}}}} \\
Subspace dimension $\tilde D$ & 16 \\
Basis $Q$ & QR decomposition of a Gaussian $D\times\tilde D$ matrix \\
Population size $\lambda$ & 5 \\
Initial step size $\sigma_0$ & 0.5 \\
Initial mean & $Q^\top x$ of the best cold start \\
Pre-conditioning & the $N_0=5$ cold starts, projected by $Q^\top$, told as the first generation \\
Update & deferred: one evaluation per step; a generation is told once $\lambda$ evaluations have accumulated \\
Lift & normalize in the subspace, map by $Q$, scale to $\bar r$ \\
CMA-ES library and options & pycma 4.4.4; population $\lambda$, verbosity $-9$, seed drawn from NumPy's global stream (seeded by the run seed) \\
\multicolumn{2}{@{}l}{\textit{Shared protocol (GuacaMol)}} \\
Tasks & 11 GuacaMol tasks \\
Seeds & 42--81 (40 seeds) \\
Cold start & $N_0$ molecules drawn without replacement from the first 10{,}000 ZINC molecules (NumPy RandomState of the seed), encoded by the encoder mean, scored directly \\
Decoding & greedy, at most 80 tokens \\
Time cap per proposal & decoding capped at 30 s, scoring not time-limited \\
Time limit per run & 3{,}600 s for both methods, a guard against hangs, not a setting of the methods \\
\bottomrule
\end{tabularx}
\end{lrbox}
\ifdim\wd\crtabbox>\linewidth\resizebox{\linewidth}{!}{\usebox{\crtabbox}}\else\usebox{\crtabbox}\fi
\end{table}

%% file: tab_sensitivity.tex
\ifdefined\crtabbox\else\newsavebox{\crtabbox}\fi
\begin{table}[t]
\centering
\caption{\textbf{Sensitivity of \geowalk{} and \geocma{} to their settings.} Top: six variants of the step schedule and the exploration of \geowalk{}, each against the default (angle $0.60\to0.15$, 16 exploration steps). Bottom: the subspace dimension $\tilde D$ of \geocma{}, 8 and 64 against the default 16. Both blocks use the same $n=440$ completed cells, as in Table~\ref{tab:main}. Means are means of per-task means. Bold marks the highest mean of the top block. The difference $\Delta$ is the variant minus the default, computed from unrounded means. The effect size $d$ is Cohen's $d$ of the cell differences. The column $p_{\mathrm{BH}}$ is the two-sided Wilcoxon signed-rank test over the cells. It is Benjamini--Hochberg corrected within each block, over 6 and 2 rows. Tasks counts the tasks on which the variant has a higher mean than the default. The column $p_{\mathrm{task}}$ is the exact two-sided Wilcoxon test on the 11 per-task mean differences, not corrected. The task-level test of the bottom block is given in the text.}
\label{tab:sensitivity}
\small
\setlength{\tabcolsep}{4pt}
\begin{lrbox}{\crtabbox}
\begin{tabularx}{\linewidth}{>{\raggedright\arraybackslash}Xrrrrrr}
\toprule
Variant & Mean & $\Delta$ & $d$ & $p_{\mathrm{BH}}$ & \begin{tabular}[b]{@{}c@{}}Tasks\\row\\higher\end{tabular} & $p_{\mathrm{task}}$ \\
\midrule
constant angle $0.60$ & $\mathbf{0.4228}$ & $+0.0001$ & $0.00$ & $0.11$ & 4/11 & $0.46$ \\
constant angle $0.35$ & $\mathbf{0.4228}$ & $+0.0001$ & $0.00$ & $0.76$ & 6/11 & $0.97$ \\
constant angle $0.15$ & $0.4189$ & $-0.0038$ & $-0.11$ & $0.006$ & 2/11 & $0.024$ \\
8 exploration steps ($\rho\approx0.32$) & $0.4202$ & $-0.0024$ & $-0.04$ & $0.76$ & 7/11 & $0.97$ \\
24 exploration steps ($\rho\approx0.96$) & $0.4133$ & $-0.0093$ & $-0.25$ & $<0.001$ & 0/11 & $0.00098$ \\
success-based angle rule (start $0.30$, $\times1.2$ / $\times0.85$, clamp $[0.05,0.50]$) & $0.4176$ & $-0.0050$ & $-0.14$ & $0.005$ & 1/11 & $0.0098$ \\
\midrule
default ($0.60\to0.15$, 16 exploration steps) (reference) & $0.4226$ & -- & -- & -- & -- & -- \\
\bottomrule
\end{tabularx}
\end{lrbox}
\ifdim\wd\crtabbox>\linewidth\resizebox{\linewidth}{!}{\usebox{\crtabbox}}\else\usebox{\crtabbox}\fi
\par\medskip
{\footnotesize\textit{\textup{\geocma{}} subspace dimension, matched set $n=440$}}\par\smallskip
\begin{lrbox}{\crtabbox}
\begin{tabularx}{\linewidth}{>{\raggedright\arraybackslash}Xrrrr}
\toprule
$\tilde D$ & Mean & $\Delta$ & $d$ & $p_{\mathrm{BH}}$ \\
\midrule
$\tilde D=16$ (default) & $0.4158$ & -- & -- & -- \\
$\tilde D=8$ & $0.4086$ & $-0.0072$ & $-0.10$ & $0.22$ \\
$\tilde D=64$ & $0.4184$ & $+0.0026$ & $0.04$ & $0.22$ \\
\bottomrule
\end{tabularx}
\end{lrbox}
\ifdim\wd\crtabbox>\linewidth\resizebox{\linewidth}{!}{\usebox{\crtabbox}}\else\usebox{\crtabbox}\fi
\end{table}

%% file: tab_pertask.tex
\ifdefined\crtabbox\else\newsavebox{\crtabbox}\fi
\begin{table}[t]
\centering
\caption{\textbf{\geowalk{} versus the GP on the shell, per task.} $\Delta$ = \geowalk{} minus the GP on the shell (Table~\ref{tab:boxsphere}). Per task, $\Delta$, Cohen's $d$ and the two-sided Wilcoxon $p$ are paired over the seeds of the task's completed cells, as in Table~\ref{tab:main}. $p_{\mathrm{Holm}}$ is Holm-corrected over the 11 tasks. Last row: all matched cells, as a mean of per-task means. The $p$ of the last row is the cell-level test over all cells. That test is not one of the per-task tests, so it has no $p_{\mathrm{Holm}}$ (--). Task codes: adip = Amlodipine MPO, dhop = Deco Hop, fexo = Fexofenadine MPO, med1 = Median molecules 1, med2 = Median molecules 2, osmb = Osimertinib MPO. Further codes: pdop = Perindopril MPO, rano = Ranolazine MPO, shop = Scaffold Hop, siga = Sitagliptin MPO, zale = Zaleplon MPO. MPO = multi-property objective.}
\label{tab:pertask}
\small
\begin{lrbox}{\crtabbox}
\begin{tabular}{lrrrrrrr}
\toprule
Task & $n$ & \geowalk{} & GP on the shell & $\Delta$ & $d$ & $p$ & $p_{\mathrm{Holm}}$ \\
\midrule
adip & 40 & $0.5013$ & $0.5037$ & $-0.0024$ & $-0.08$ & $0.92$ & $1.00$ \\
dhop & 40 & $0.5706$ & $0.5713$ & $-0.0007$ & $-0.05$ & $0.82$ & $1.00$ \\
fexo & 40 & $0.6760$ & $0.6753$ & $+0.0006$ & $+0.01$ & $0.83$ & $1.00$ \\
med1 & 40 & $0.1287$ & $0.1221$ & $+0.0066$ & $+0.40$ & $0.024$ & $0.27$ \\
med2 & 40 & $0.1525$ & $0.1512$ & $+0.0013$ & $+0.13$ & $0.33$ & $1.00$ \\
osmb & 40 & $0.7142$ & $0.7254$ & $-0.0112$ & $-0.21$ & $0.44$ & $1.00$ \\
pdop & 40 & $0.4838$ & $0.4769$ & $+0.0069$ & $+0.13$ & $0.47$ & $1.00$ \\
rano & 40 & $0.5393$ & $0.5322$ & $+0.0071$ & $+0.08$ & $0.72$ & $1.00$ \\
shop & 40 & $0.4400$ & $0.4406$ & $-0.0005$ & $-0.03$ & $0.82$ & $1.00$ \\
siga & 40 & $0.0988$ & $0.1123$ & $-0.0135$ & $-0.14$ & $0.78$ & $1.00$ \\
zale & 40 & $0.3438$ & $0.3393$ & $+0.0044$ & $+0.04$ & $0.97$ & $1.00$ \\
\midrule
all tasks & 440 & $0.4226$ & $0.4228$ & $-0.0001$ & $0.00$ & $0.50$ & -- \\
\bottomrule
\end{tabular}
\end{lrbox}
\ifdim\wd\crtabbox>\linewidth\resizebox{\linewidth}{!}{\usebox{\crtabbox}}\else\usebox{\crtabbox}\fi
\end{table}

%% file: tab_pertask_full.tex
\ifdefined\crtabbox\else\newsavebox{\crtabbox}\fi
\begin{table}[t]
\centering
\caption{\textbf{Per-task means at $N=30$.} All columns use one matched set of $n=440$ completed cells, as in Table~\ref{tab:main}. Task codes as in Table~\ref{tab:pertask}. Last row: mean of per-task means. Bold: best per task. Box = $[p_{01},p_{99}]$. The GP on the shell and the GP in the box are the two DSP rows of Table~\ref{tab:boxsphere}; the Linear BO ports are those of Table~\ref{tab:trajectories}.}
\label{tab:pertask_full}
\footnotesize
\setlength{\tabcolsep}{4pt}
\begin{lrbox}{\crtabbox}
\begin{tabular}{lrrrrrrrr}
\toprule
Task & $n$ & \geowalk{} & \geocma{} & \begin{tabular}[b]{@{}c@{}}Random Search\\(shell)\end{tabular} & \begin{tabular}[b]{@{}c@{}}GP on\\the shell\end{tabular} & \begin{tabular}[b]{@{}c@{}}GP in\\the box\end{tabular} & \begin{tabular}[b]{@{}c@{}}Linear BO port\\(shell)\end{tabular} & \begin{tabular}[b]{@{}c@{}}Linear BO port\\(box)\end{tabular} \\
\midrule
adip & 40 & $0.501$ & $0.499$ & $0.498$ & $\mathbf{0.504}$ & $0.459$ & $0.500$ & $0.474$ \\
dhop & 40 & $0.571$ & $0.569$ & $0.568$ & $\mathbf{0.571}$ & $0.561$ & $0.570$ & $0.560$ \\
fexo & 40 & $\mathbf{0.676}$ & $0.669$ & $0.662$ & $0.675$ & $0.628$ & $0.666$ & $0.628$ \\
med1 & 40 & $0.129$ & $0.121$ & $0.120$ & $0.122$ & $0.154$ & $0.121$ & $\mathbf{0.156}$ \\
med2 & 40 & $0.153$ & $0.152$ & $0.151$ & $0.151$ & $0.150$ & $\mathbf{0.153}$ & $0.150$ \\
osmb & 40 & $0.714$ & $0.719$ & $0.708$ & $\mathbf{0.725}$ & $0.702$ & $0.719$ & $0.714$ \\
pdop & 40 & $\mathbf{0.484}$ & $0.463$ & $0.469$ & $0.477$ & $0.378$ & $0.475$ & $0.405$ \\
rano & 40 & $\mathbf{0.539}$ & $0.516$ & $0.507$ & $0.532$ & $0.477$ & $0.531$ & $0.502$ \\
shop & 40 & $0.440$ & $0.440$ & $0.436$ & $\mathbf{0.441}$ & $0.427$ & $0.438$ & $0.424$ \\
siga & 40 & $0.099$ & $0.076$ & $0.089$ & $0.112$ & $\mathbf{0.128}$ & $0.098$ & $0.125$ \\
zale & 40 & $0.344$ & $\mathbf{0.349}$ & $0.323$ & $0.339$ & $0.323$ & $0.340$ & $0.349$ \\
\midrule
mean & 440 & $0.423$ & $0.416$ & $0.412$ & $\mathbf{0.423}$ & $0.399$ & $0.419$ & $0.408$ \\
\bottomrule
\end{tabular}
\end{lrbox}
\ifdim\wd\crtabbox>\linewidth\resizebox{\linewidth}{!}{\usebox{\crtabbox}}\else\usebox{\crtabbox}\fi
\end{table}

%% file: tab_trajectories.tex
\ifdefined\crtabbox\else\newsavebox{\crtabbox}\fi
\begin{table}[t]
\centering
\caption{\textbf{Best-so-far after $N=10$, $20$ and $30$ evaluations on one matched set.} All rows except the one marked $^\ddagger$ use the same $n=440$ completed cells, as in Table~\ref{tab:main}. The row marked $^\ddagger$ covers $n=439$ of them (one run did not complete). Values are means of per-task means of the best score after $N$ evaluations: 5 shared cold-start molecules and $N-5$ proposals. Bold marks the highest mean in the column. Linear BO is the published recipe of Table~\ref{tab:published}; the two ports are ours (Appendix~\ref{app:baselines}). Bottom block: \geowalk{} minus the GP on the shell, paired on the same cells. The effect size $d$ is Cohen's $d$ of the cell differences. The column $p_{\mathrm{BH}}$ is the two-sided Wilcoxon signed-rank test over the cells. It is Benjamini--Hochberg corrected over the 3 budgets $N=10$, $20$ and $30$.}
\label{tab:trajectories}
\footnotesize
\setlength{\tabcolsep}{4pt}
\begin{lrbox}{\crtabbox}
\begin{tabularx}{\linewidth}{>{\raggedright\arraybackslash}X>{\raggedright\arraybackslash}p{2.2cm}rrr}
\toprule
Method & Support & $N=10$ & $N=20$ & $N=30$ \\
\midrule
\geowalk{} & shell & $0.3612$ & $0.3978$ & $0.4226$ \\
GP on the shell & shell & $0.3673$ & $\mathbf{0.4067}$ & $\mathbf{0.4228}$ \\
Linear BO port to the shell (calibrated bounds) & shell & $\mathbf{0.3681}$ & $0.4066$ & $0.4192$ \\
Random Search (shell) & shell & $0.3612$ & $0.3978$ & $0.4118$ \\
Linear BO port, box & $[p_{01},p_{99}]$ & $0.3499$ & $0.3887$ & $0.4079$ \\
Linear BO & box $[-8,8]^D$ & $0.3479$ & $0.3907$ & $0.4101$ \\
GP in the box & $[p_{01},p_{99}]$ & $0.3435$ & $0.3784$ & $0.3989$ \\
TuRBO & trust region in $[p_{01},p_{99}]$ & $0.3350$ & $0.3607$ & $0.3790$ \\
BAxUS$^\ddagger$ & nested subspaces of $[p_{01},p_{99}]$ & $0.3336$ & $0.3662$ & $0.3907$ \\
\bottomrule
\end{tabularx}
\end{lrbox}
\ifdim\wd\crtabbox>\linewidth\resizebox{\linewidth}{!}{\usebox{\crtabbox}}\else\usebox{\crtabbox}\fi
\par\medskip
{\footnotesize\textit{\textup{\geowalk{}} $-$ the GP on the shell, paired on the same cells}}\par\smallskip
\begin{lrbox}{\crtabbox}
\begin{tabularx}{\linewidth}{>{\raggedright\arraybackslash}Xrrr}
\toprule
 & $N=10$ & $N=20$ & $N=30$ \\
\midrule
$\Delta$ & $-0.0062$ & $-0.0090$ & $-0.0001$ \\
$d$ & $-0.07$ & $-0.13$ & $0.00$ \\
$p_{\mathrm{BH}}$ & $0.27$ & $0.002$ & $0.50$ \\
Tasks, \geowalk{} higher & 2/11 & 0/11 & 6/11 \\
\bottomrule
\end{tabularx}
\end{lrbox}
\ifdim\wd\crtabbox>\linewidth\resizebox{\linewidth}{!}{\usebox{\crtabbox}}\else\usebox{\crtabbox}\fi
\end{table}

%% file: tab_crosscodec.tex
\ifdefined\crtabbox\else\newsavebox{\crtabbox}\fi
\begin{table}[t]
\centering
\caption{\textbf{Scope of the codec comparison (Appendix~\ref{app:crosscodec}).} The GP on the shell uses candidate-set argmax and the GP in the box L-BFGS-B, so shell versus box also changes the maximizer. Each codec has its own box $[p_{01},p_{99}]$ in its own latent space. Each codec row uses the matched cells of its four methods (completed cells, as in Table~\ref{tab:main}). Means are means of per-task means. $\Delta_{\mathrm{shell-box}}$ = the GP on the shell minus the GP in the box. $d$ = Cohen's $d$ of the cell differences. $p$ = two-sided Wilcoxon signed-rank test over the cells, not corrected. The task-level test is given in the text.}
\label{tab:crosscodec_means}
\footnotesize
\setlength{\tabcolsep}{3pt}
\begin{lrbox}{\crtabbox}
\begin{tabularx}{\linewidth}{>{\raggedright\arraybackslash}X>{\raggedright\arraybackslash}p{1.5cm}rrrrrrrr}
\toprule
Codec & $z$ normalized in decoder; decoding & $n$ & \geowalk{} & \geocma{} & \begin{tabular}[b]{@{}c@{}}GP on\\the shell\end{tabular} & \begin{tabular}[b]{@{}c@{}}GP in\\the box\end{tabular} & $\Delta_{\mathrm{shell-box}}$ & $d$ & $p$ \\
\midrule
SELFIES VAE & no, deterministic & 440 & $0.4226$ & $0.4158$ & $0.4228$ & $0.3989$ & $+0.0239$ & $0.30$ & $<0.001$ \\
Chembed (Transformer VAE) & no, deterministic & 440 & $0.3765$ & $0.3734$ & $0.3860$ & $0.3810$ & $+0.0050$ & $0.06$ & $0.26$ \\
Power-Spherical VAE & yes, stochastic & 440 & $0.3739$ & $0.3706$ & $0.3689$ & $0.3594$ & $+0.0094$ & $0.11$ & $0.052$ \\
\bottomrule
\end{tabularx}
\end{lrbox}
\ifdim\wd\crtabbox>\linewidth\resizebox{\linewidth}{!}{\usebox{\crtabbox}}\else\usebox{\crtabbox}\fi
\end{table}

%% file: tab_budget.tex
\ifdefined\crtabbox\else\newsavebox{\crtabbox}\fi
\begin{table}[t]
\centering
\caption{\textbf{Best-so-far of the stored $N=1000$ runs at intermediate budgets (seeds 42--51).} One matched set of $n=109$ of the 110 (task, seed) cells. Each column is the best value after $N$ evaluations of the same runs. The $N=30$ column is therefore a prefix of these runs, not the $N=30$ results of Table~\ref{tab:main}. GP in the cold-start box: a Thompson-sampling GP in the bounding box of the five cold-start latents, not the box version of Table~\ref{tab:boxsphere}. Bold: the highest mean in the column. The $\Delta$ rows are paired differences on the matched cells.}
\label{tab:budget}
\small
\setlength{\tabcolsep}{4pt}
\begin{lrbox}{\crtabbox}
\begin{tabularx}{\linewidth}{>{\raggedright\arraybackslash}Xrrrrrr}
\toprule
Method & $N=30$ & $N=50$ & $N=100$ & $N=200$ & $N=500$ & $N=1000$ \\
\midrule
\geowalk{} & $0.424$ & $\mathbf{0.445}$ & $\mathbf{0.463}$ & $\mathbf{0.479}$ & $\mathbf{0.497}$ & $\mathbf{0.509}$ \\
GP on the shell & $\mathbf{0.426}$ & $0.443$ & $0.459$ & $0.471$ & $0.489$ & $0.501$ \\
\geocma{} & $0.414$ & $0.439$ & $0.461$ & $0.474$ & $0.487$ & $0.495$ \\
Random Search (shell) & $0.409$ & $0.424$ & $0.444$ & $0.460$ & $0.478$ & $0.491$ \\
GP in the cold-start box & $0.380$ & $0.395$ & $0.420$ & $0.430$ & $0.446$ & $0.458$ \\
\midrule
$\Delta$ \geowalk{} $-$ GP on the shell & $-0.002$ & $+0.002$ & $+0.004$ & $+0.008$ & $+0.008$ & $+0.008$ \\
$\Delta$ \geowalk{} $-$ GP in the cold-start box & $+0.044$ & $+0.050$ & $+0.043$ & $+0.049$ & $+0.051$ & $+0.051$ \\
\bottomrule
\end{tabularx}
\end{lrbox}
\ifdim\wd\crtabbox>\linewidth\resizebox{\linewidth}{!}{\usebox{\crtabbox}}\else\usebox{\crtabbox}\fi
\end{table}

%% file: tab_lassobench.tex
\ifdefined\crtabbox\else\newsavebox{\crtabbox}\fi
\begin{table}[t]
\centering
\caption{\textbf{LassoBench: final loss divided by the task's random-search mean loss, per dimension.} Each value is the mean over seeds of this ratio after 30 evaluations. A value of 1 equals random search, and a value above 1 is worse than random search. Lower is better. The tasks are $D=60$ synthetic simple, $D=100$ synthetic medium, $D=180$ DNA, $D=300$ synthetic high, $D=1000$ synthetic hard, $D=7129$ leukemia. In parentheses: the seeds with a finite loss ($n/40$ where fewer than the 40 seeds 42--81 have one). n.c.: no seed finished with a finite loss. n.a.: not applicable at this dimension, so not run. The coverage and time-outs of each method, and each method minus \geocma{}, are in the full table in the repository.}
\label{tab:lassobench}
\scriptsize
\setlength{\tabcolsep}{3pt}
\begin{lrbox}{\crtabbox}
\begin{tabularx}{\linewidth}{>{\raggedright\arraybackslash}Xrrrrrr}
\toprule
Method & $D=60$ & $D=100$ & $D=180$ & $D=300$ & $D=1000$ & $D=7129$ \\
\midrule
\geocma{} & $0.805$\,(40) & $0.351$\,(40) & $0.891$\,(40) & $0.126$\,(40) & $0.127$\,(40) & $0.672$\,(40) \\
\geowalk{} & $0.566$\,(40) & $0.687$\,(40) & $0.974$\,(40) & $0.748$\,(40) & $0.868$\,(40) & $0.931$\,(40) \\
Random search & $1.000$\,(40) & $1.000$\,(40) & $1.000$\,(40) & $1.000$\,(40) & $1.000$\,(40) & $1.000$\,(40) \\
RBF GP (DSP prior) & $2.160$\,(40) & $1.246$\,(40) & $0.968$\,(40) & $0.687$\,(40) & $0.676$\,(40) & $0.869$\,(40) \\
Mat\'ern-5/2 GP ($\ell_0=0.1\sqrt{D}$) & $1.141$\,(40) & $0.793$\,(40) & $0.994$\,(40) & $0.890$\,(38/40) & $0.865$\,(40) & $0.916$\,(36/40) \\
Mat\'ern-5/2 GP ($\ell_0=\sqrt{D}$) & $0.758$\,(40) & $0.492$\,(40) & $1.004$\,(39/40) & $0.730$\,(25/40) & $0.711$\,(38/40) & n.c. \\
Linear BO port & $0.879$\,(40) & $0.243$\,(40) & $0.896$\,(40) & $0.138$\,(40) & $0.133$\,(40) & n.c. \\
\multicolumn{7}{@{}l}{\textit{Published methods}} \\
Vanilla BO & $1.710$\,(40) & $0.856$\,(40) & $0.954$\,(40) & $0.754$\,(40) & $0.685$\,(40) & $0.899$\,(40) \\
Papenmeier MSR & $0.692$\,(40) & $1.011$\,(40) & $0.994$\,(40) & $0.819$\,(40) & $0.717$\,(17/40) & n.c. \\
Papenmeier MLE (scaled) & $0.616$\,(40) & $0.519$\,(40) & $0.998$\,(40) & $0.834$\,(40) & $0.946$\,(29/40) & n.c. \\
Xu Standard GP & $0.400$\,(40) & $0.356$\,(40) & $1.016$\,(40) & $0.806$\,(40) & n.a. & n.a. \\
Linear BO & $0.759$\,(40) & $0.271$\,(40) & $0.898$\,(40) & $0.125$\,(40) & $0.131$\,(40) & n.c. \\
TuRBO & $1.298$\,(40) & $0.789$\,(40) & $0.962$\,(40) & $0.759$\,(40) & $0.963$\,(40) & $1.148$\,(40) \\
BAxUS & $0.997$\,(40) & $0.399$\,(40) & $0.873$\,(40) & $0.115$\,(40) & $0.124$\,(40) & $0.709$\,(40) \\
CMA-ES (LassoBench recipe) & $0.478$\,(40) & $0.287$\,(40) & $0.874$\,(40) & $0.109$\,(40) & $0.125$\,(40) & $0.692$\,(40) \\
\bottomrule
\end{tabularx}
\end{lrbox}
\ifdim\wd\crtabbox>\linewidth\resizebox{\linewidth}{!}{\usebox{\crtabbox}}\else\usebox{\crtabbox}\fi
\end{table}
\begin{table}[t]
\centering
\caption{\textbf{LassoBench, continued from Table~\ref{tab:lassobench}: \geowalk{} against each published method.} Same runs and normalization as Table~\ref{tab:lassobench}. These methods form the baseline family of LassoBench. Here a task is a dimension. The difference $\Delta$ is \geowalk{} minus the method, a mean over tasks of the per-task mean normalized losses. A positive $\Delta$ means that \geowalk{} has the higher, worse loss. Tasks counts the tasks on which \geowalk{} has the lower mean loss, of those where both have runs. The column $n$ gives the (task, seed) pairs that both completed. The column $p_{\mathrm{task}}$ is the exact two-sided Wilcoxon test on the per-task mean differences. The column $p_{\mathrm{task}}^{\mathrm{BH}}$ is $p_{\mathrm{task}}$ corrected by Benjamini--Hochberg over the 8 methods. Bold marks values below 0.05. With $k$ tasks the exact two-sided test cannot go below $2/2^k$, so a method with fewer than 6 tasks cannot reach 0.05. The mark $^{\|}$ denotes the two arms that ran only on LassoBench; Table~\ref{tab:main} prints the same method names under other settings. This family has 8 methods, and the family of Table~\ref{tab:published} has 6 recipes.}
\label{tab:lassobench_family}
\scriptsize
\setlength{\tabcolsep}{3pt}
{\footnotesize\textit{\textup{\geowalk{}} against each published method}}\par\smallskip
\begin{lrbox}{\crtabbox}
\begin{tabularx}{\linewidth}{>{\raggedright\arraybackslash}Xrrrrr}
\toprule
Method & \begin{tabular}[b]{@{}c@{}}$\Delta$\\(\geowalk{} $-$ method)\end{tabular} & \begin{tabular}[b]{@{}c@{}}Tasks\\\geowalk{} lower\end{tabular} & $n$ & $p_{\mathrm{task}}$ & $p_{\mathrm{task}}^{\mathrm{BH}}$ \\
\midrule
Vanilla BO & $-0.181$ & 3/6 & 240 & $1.00$ & $1.00$ \\
Papenmeier MSR & $-0.077$ & 4/5 & 177 & $0.44$ & $0.70$ \\
Papenmeier MLE (scaled) & $-0.012$ & 4/5 & 189 & $0.62$ & $0.71$ \\
Xu Standard GP & $+0.099$ & 2/4 & 160 & $0.62$ & $0.71$ \\
Linear BO & $+0.332$ & 1/5 & 200 & $0.19$ & $0.44$ \\
TuRBO & $-0.191$ & 5/6 & 240 & $0.094$ & $0.38$ \\
BAxUS$^{\|}$ & $+0.259$ & 1/6 & 240 & $0.22$ & $0.44$ \\
CMA-ES (LassoBench recipe)$^{\|}$ & $+0.368$ & 0/6 & 240 & $0.031$ & $0.25$ \\
\bottomrule
\end{tabularx}
\end{lrbox}
\ifdim\wd\crtabbox>\linewidth\resizebox{\linewidth}{!}{\usebox{\crtabbox}}\else\usebox{\crtabbox}\fi
\end{table}

%% file: tab_main_notes.tex
\paragraph{Notes to Table~\ref{tab:main}.} All rows are evaluated on one matched set of $n=440$ (task, seed) cells. It is the intersection of the completed cells of the rows of the full version of the table, except the rows that cover only part of it. A completed cell is a run that made all 25 proposals after the 5 cold starts. In every row printed here, such a run reached all 30 evaluations. Runs in which a surrogate fit failed and a step fell back to a uniform proposal are included. The row marked $^\ddagger$ covers only part of that set and uses the part it covers (BAxUS: $n=439$, one run did not complete). Cross-mean = mean of per-task means, and bold marks the highest. The difference $\Delta$ is \geowalk{} minus the row on the same cells, as a mean of per-task means of the paired differences. W/L counts the cells won and lost by \geowalk{}. A cell with $|\mathrm{diff}|\le10^{-6}$ is a tie and is excluded from W/L. The effect size $d$ is the mean of the cell differences divided by their standard deviation. The column $p_{\mathrm{task}}$ is the exact two-sided Wilcoxon test on the per-task mean differences. It is the primary test of the paper and is printed here without correction. The full version of this table is released with the code (\texttt{docs/baselines/TABLE1\_FULL.md} and \texttt{docs/baselines/tab\_main\_full.tex}). It has all 31 comparator rows under the names of their implementation, and the cell-level Wilcoxon test, Benjamini--Hochberg corrected over the 31 rows. The rows not printed here are in the full version. Table~\ref{tab:boxsphere} compares the GP recipes on the box and on the shell. TuRBO, Vanilla BO, Papenmeier MSR, Xu Standard GP and Linear BO are the published recipes, rebuilt from the authors' released code; Table~\ref{tab:published} compares them with \geowalk{} and the GP on the shell. The other rows adapt the cited method to this benchmark. Their deviations from the published methods are summarized in Table~\ref{tab:baselines} and listed in full in the method cards released with the code (\texttt{docs/baselines/}). The full version names each row by its implementation: CMA-ES is ``CMA-ES in $\mathbb{R}^D$, $\sigma_0=0.5$, population 5''; BAxUS is ``GP in nested count-sketch subspaces, Thompson sampling in a trust region''; REMBO is ``GP in a random linear embedding, Thompson sampling''; HeSBO is ``GP in a hashed embedding, Thompson sampling''; GaBO is ``sphere Gaussian-kernel GP, EI, Riemannian trust-region maximizer''; Riemannian Mat\'ern GP is ``Riemannian Mat\'ern GP, EI, Riemannian trust-region maximizer''. The GP on the shell of Section~\ref{sec:main_results}, which Table~\ref{tab:boxsphere} prints with its box version, is ``Mat\'ern-5/2 GP (DSP prior), qLogEI, candidate-set argmax''. Box rows use the $[p_{01},p_{99}]$ box, except Linear BO, which runs in $[-8,8]^D$. Table~\ref{tab:baselines} gives the support of every row. Abbreviations: DSP prior is the dimension-scaled lengthscale prior $\mathrm{LogNormal}(\sqrt{2}+0.5\ln D,\sqrt{3})$ of Hvarfner et al. The value $\ell_0$ is the initial lengthscale before the maximum-likelihood fit. EI, LogEI and qLogEI are expected improvement, its logarithm (analytic) and its Monte Carlo batch version. Candidate-set argmax: the acquisition is maximized over a finite set of random candidates on the support. Sample-around-best: L-BFGS-B restarts include points sampled around the best observations. MSR and MLE (scaled) are the two recipes of Papenmeier et al., with maximum-likelihood lengthscales initialized at $0.1\sqrt{D}$. MSR adds RAASP perturbations (random axis-aligned subspace perturbations) in the acquisition maximizer, and MLE (scaled) does not. The $[p_{01},p_{99}]$ box spans the per-coordinate 1st to 99th percentile of encoded ZINC latents. The radius $\bar r$ of the shell is the mean norm of the run's five cold-start latents. The set $S^{D-1}$ is the unit sphere.

%% file: tab_baselines_overview.tex
{\scriptsize\setlength{\tabcolsep}{2pt}
\begin{longtable}{@{}>{\raggedright\arraybackslash}p{2.7cm}>{\raggedright\arraybackslash}p{2.0cm}>{\raggedright\arraybackslash}p{2.4cm}>{\raggedright\arraybackslash}p{2.4cm}>{\raggedright\arraybackslash}p{2.4cm}@{}}
\caption{\textbf{Overview of every printed method, one row per method and its variants.} Implementation: where the code comes from. Surrogate: model and fit. Acquisition, maximizer: how a proposal is chosen. Support, trust region: where the candidates are drawn ($\bar r$: mean norm of the run's five cold-start latents). Abbreviations: ARD = one lengthscale per input dimension. MAP = maximum a posteriori fit of the hyperparameters. RAASP = random axis-aligned subspace perturbations. qLogNEI = Monte Carlo log noisy expected improvement. UCB = upper confidence bound. PCA sphere = the sphere of radius $\bar r$ in the span of the top 64 principal components of 3,000 encoded ZINC latents. The two BAxUS rows run the BoTorch tutorial, not the authors' code, so Table~\ref{tab:main} counts that method as an adapted one. Our adapted Mat\'ern-5/2 GP uses the DSP prior with location $\sqrt{2}+0.5\ln D$, as in the paper of Hvarfner et al. Their released configuration, which the published recipe follows, uses 1.4 in place of $\sqrt{2}$. GaBO fell back to a uniform draw in \crBaselinesGaboFallbackSteps{} proposals, in \crBaselinesGaboFallbackCells{} of its \crBaselinesGaboCells{} runs. The full card of every method and its run status are released with the code (\texttt{docs/baselines/}).}\label{tab:baselines}\\
\toprule
Method & Implementation & Surrogate & Acquisition, maximizer & Support, trust region \\
\midrule
\endfirsthead
\toprule
Method & Implementation & Surrogate & Acquisition, maximizer & Support, trust region \\
\midrule
\endhead
\bottomrule
\endfoot
\multicolumn{5}{@{}l}{\textit{Methods of this paper (GuacaMol, LassoBench and NAS-Bench-101)}} \\
\geowalk{} and its variants & this paper & none & 16 uniform draws, then geodesic steps, angle 0.60 to 0.15 rad & shell of radius $\bar r$ (NAS: mean embedding norm; LassoBench: clipped) \\
\geocma{} and its variants & this paper; CMA-ES from pycma & none (CMA-ES) & 16-dim. subspace, population 5, step 0.5 (NAS: sd of cold starts) & shell of radius $\bar r$ (NAS: whole space, mean norm; LassoBench: box) \\
Random search, on the shell and on other supports & this paper & none & uniform draws; Gaussian fit: $N(\hat\mu,\hat\Sigma)$ draws; NAS: 30 table entries & shell, PCA sphere, convex combinations of latents, Gaussian; LassoBench: $[-1,1]^D$ \\
\multicolumn{5}{@{}l}{\textit{GP recipes adapted in this repository}} \\
Mat\'ern-5/2 GP (DSP prior) & own BoTorch GP; adapts Hvarfner et al. 2024 & Mat\'ern-5/2, ARD, DSP lengthscale prior, MAP fit & qLogEI; argmax of 2,048 draws or L-BFGS-B & shell, box $[p_{01},p_{99}]$, or cube directions at $\bar r$ \\
Mat\'ern-5/2 GP ($\ell_0=0.1\sqrt{D}$) & own BoTorch GP; adapts Papenmeier et al. 2025 & Mat\'ern-5/2, ARD, start $0.1\sqrt{D}$, no lengthscale prior & qLogEI, argmax or L-BFGS-B; one row LogEI with RAASP & shell or box $[p_{01},p_{99}]$; LassoBench: $[-1,1]^D$ \\
Mat\'ern-5/2 GP ($\ell_0=\sqrt{D}$) & own BoTorch GP; adapts Xu et al. 2025 & Mat\'ern-5/2, ARD, start $\sqrt{D}$; no lengthscale/\allowbreak{}outputscale prior & qLogEI; argmax over 2,048 candidates or L-BFGS-B & shell or box $[p_{01},p_{99}]$; LassoBench: $[-1,1]^D$ \\
Linear BO port, box & kernel copied from Doumont et al. & spherical linear kernel, LogNormal($\sqrt{2}$, $\sqrt{3}$) prior & LogEI; L-BFGS-B, 4 restarts, sample-around-best & box $[p_{01},p_{99}]$ (rescaled); LassoBench: $[-1,1]^D$ \\
Linear BO port to the shell (calibrated bounds) & kernel copied from Doumont et al. & as the box port; kernel bounds calibrated to the shell & qLogEI; argmax over 2,048 candidates on the shell & shell; the kernel-bound rule was set after a first run on the shell \\
RBF GP (DSP prior) & own BoTorch GP; adapts Hvarfner et al. 2024 & RBF, ARD, LogNormal(1.4 + 0.5 $\ln D$, $\sqrt{3}$), MAP & qLogNEI; L-BFGS-B, 4 restarts, sample-around-best & $[-1,1]^D$ (LassoBench only) \\
\multicolumn{5}{@{}l}{\textit{Published recipes}} \\
Vanilla BO & authors' code (Ax 0.3.7) rebuilt on BoTorch & RBF, ARD, LogNormal(1.4 + 0.5 $\ln D$, $\sqrt{3}$), MAP & qLogNEI (in-box baseline); BoTorch, 4 restarts & box $[p_{01},p_{99}]$; LassoBench: $[-1,1]^D$ \\
Vanilla BO, all observations as baseline (the other run of the same recipe) & as the row above & as the row above & qLogNEI, all observations as baseline; BoTorch & box $[p_{01},p_{99}]$ \\
Papenmeier MSR & authors' released code, rebuilt & Mat\'ern-5/2, ARD, start $\sqrt{D}/10$, learned noise & LogEI; 5 restarts, RAASP (sigma 0.001) & box $[p_{01},p_{99}]$; LassoBench: $[-1,1]^D$; no trust region \\
Papenmeier MLE (scaled) & authors' released code, rebuilt & Mat\'ern-5/2, ARD, start $\sqrt{D}/10$, learned noise & LogEI; 5 restarts, no sampling around the best & box $[p_{01},p_{99}]$; LassoBench: $[-1,1]^D$ \\
Xu Standard GP & authors' released code, rebuilt & Mat\'ern-5/2, ARD, Uniform(1e-10, 30) prior, start $\sqrt{D}$ & UCB (beta 1.5); 10 restarts, 1,000 raw samples & box $[p_{01},p_{99}]$; LassoBench: $[-1,1]^D$, not applicable (n.a.) on the two largest tasks \\
Linear BO & authors' released code; kernel copied & spherical linear kernel, LogNormal($\sqrt{2}$, $\sqrt{3}$) prior & LogEI; L-BFGS-B, 4 restarts, sample-around-best & authors' box $[-8,8]^{256}$; LassoBench: $[-1,1]^D$ \\
TuRBO & authors' turbo\_1.py and gp.py, ported & Mat\'ern-5/2, ARD, ExactGP trained by Adam (50 steps) & Thompson sampling of the predictive over Sobol candidates & TuRBO-1 trust region in $[p_{01},p_{99}]$; LassoBench: $[-1,1]^D$ \\
BAxUS (LassoBench) & BAxUS tutorial of BoTorch (tutorials/baxus) & BoTorch default RBF GP on the embedding & Thompson sampling; argmax of one posterior sample & nested count-sketch embeddings, BAxUS trust region \\
BAxUS (GuacaMol) & BAxUS tutorial of BoTorch (tutorials/baxus) & BoTorch default RBF GP on the embedding & Thompson sampling; argmax of one posterior sample & GuacaMol: nested count-sketch embeddings of the box $[p_{01},p_{99}]$, BAxUS trust region \\
CMA-ES (LassoBench recipe) & LassoBench's CMA-ES example & none (pycma 4.4.4) & CMA-ES, population 20, step size 0.1, $x_0=0$ & $[-1,1]^D$ (LassoBench only) \\
\multicolumn{5}{@{}l}{\textit{Other baselines implemented in this repository}} \\
REMBO and HeSBO & re-implements REMBO (Wang et al.), HeSBO (Nayebi et al.) & BoTorch default RBF GP on the 16-dim. embedding & Thompson sampling; argmax over 100 points of the box & 16-dim. embedding lifted to the shell \\
GaBO & vendored GaBOtorch (Jaquier et al.) at 884f64a & sphere Gaussian kernel, Gamma(2, 0.15) outputscale prior & EI; Riemannian trust regions; uniform draw on failure & unit sphere $S^{255}$, proposals scaled to $\bar r$ \\
Riemannian Mat\'ern GP & vendored MaternGaBO (Jaquier et al.) at 84bffda & Riemannian Mat\'ern kernel on the sphere, nu = 2.5 & EI; robust Riemannian trust regions, 5 restarts & unit sphere $S^{255}$, proposals scaled to $\bar r$ \\
CMA-ES (population 5 or 20) & this repository; CMA-ES from pycma & none (CMA-ES) & CMA-ES in $\mathbb{R}^{256}$, step size 0.5, population 5 or 20 & $\mathbb{R}^{256}$, no bounds \\
\multicolumn{5}{@{}l}{\textit{GuacaMol stored runs}} \\
GP in the cold-start box ($N=1000$ runs) & this repository; not the recipe of Hvarfner et al. & Mat\'ern-5/2, ARD, gpytorch defaults, no explicit priors & Thompson sampling; argmax over 2,560 Sobol points & bounding box of the five cold-start encodings \\
\end{longtable}}

%% file: tab_published.tex
\ifdefined\crtabbox\else\newsavebox{\crtabbox}\fi
\begin{table}[t]
\centering
\caption{\textbf{Published recipes against \geowalk{} and the GP on the shell at $N=30$ (baseline family).} All rows use one matched set of $n=440$ cells. Mean and $\Delta$ are means of per-task means of best@30, and $\Delta$ = the reference minus the recipe. The reference is \geowalk{} in the left group of columns and the GP on the shell in the right group. Tasks = tasks, of the 11, on which the reference has the higher mean. $p_{\mathrm{task}}$ = exact two-sided Wilcoxon test on the 11 per-task differences, the primary test. It is corrected by Benjamini--Hochberg within the 12 tests of this table. Bold marks $p_{\mathrm{task}}^{\mathrm{BH}}<0.05$. The last row is the other run of the Hvarfner recipe, outside the corrected family; Table~\ref{tab:baselines} describes both runs.}
\label{tab:published}
\footnotesize
\setlength{\tabcolsep}{3pt}
\begin{lrbox}{\crtabbox}
\begin{tabularx}{\linewidth}{>{\raggedright\arraybackslash}Xrrrrrrrrr}
\toprule
\multicolumn{2}{c}{} & \multicolumn{4}{c}{Reference: \geowalk{}} & \multicolumn{4}{c}{Reference: GP on the shell} \\
\cmidrule(lr){3-6}\cmidrule(lr){7-10}
Recipe & Mean & $\Delta$ & Tasks & $p_{\mathrm{task}}$ & $p_{\mathrm{task}}^{\mathrm{BH}}$ & $\Delta$ & Tasks & $p_{\mathrm{task}}$ & $p_{\mathrm{task}}^{\mathrm{BH}}$ \\
\midrule
Vanilla BO & $0.4008$ & $+0.0218$ & 9/11 & $0.083$ & $0.091$ & $+0.0220$ & 7/11 & $0.15$ & $0.15$ \\
Papenmeier MSR & $0.3899$ & $+0.0327$ & 9/11 & $0.019$ & $0.056$ & $+0.0329$ & 10/11 & $0.0098$ & $\mathbf{0.039}$ \\
Papenmeier MLE (scaled) & $0.3892$ & $+0.0334$ & 9/11 & $0.0098$ & $\mathbf{0.039}$ & $+0.0335$ & 10/11 & $0.0098$ & $\mathbf{0.039}$ \\
Xu Standard GP & $0.3951$ & $+0.0275$ & 9/11 & $0.042$ & $0.063$ & $+0.0277$ & 9/11 & $0.054$ & $0.064$ \\
Linear BO & $0.4101$ & $+0.0125$ & 10/11 & $0.032$ & $0.063$ & $+0.0127$ & 9/11 & $0.054$ & $0.064$ \\
TuRBO & $0.3790$ & $+0.0437$ & 9/11 & $0.032$ & $0.063$ & $+0.0438$ & 8/11 & $0.042$ & $0.063$ \\
\midrule
Vanilla BO, all observations as baseline (the other run of the same recipe) & $0.3993$ & $+0.0233$ & 8/11 & $0.12$ & -- & $+0.0234$ & 8/11 & $0.10$ & -- \\
\bottomrule
\end{tabularx}
\end{lrbox}
\ifdim\wd\crtabbox>\linewidth\resizebox{\linewidth}{!}{\usebox{\crtabbox}}\else\usebox{\crtabbox}\fi
\end{table}

%% file: tab_geometry.tex
\ifdefined\crtabbox\else\newsavebox{\crtabbox}\fi
\begin{table}[t]
\centering
\caption{\textbf{Shell geometry of the encoder means $\mu(x)$ of the VAE ($D=256$), on ZINC-250k.} $\Sigma$ is the covariance of $\mu(x)$ and $m=\mathrm{tr}\,\Sigma+\|\bar\mu\|^2$, both with the $n-1$ normalization. SD is the standard deviation of $\|\mu\|$, also with $n-1$. The model radius and width are second-order (Gaussian) delta-method values, with $\mathrm{Var}(\|\mu\|^2)=2\,\mathrm{tr}(\Sigma^2)+4\bar\mu^\top\Sigma\bar\mu$. They are computed on the same encodings, so they describe the shell and do not predict it. Dropped tokens: SELFIES tokens outside the vocabulary of the VAE (stereo tokens) are removed before encoding.}
\label{tab:geometry}
\footnotesize
\setlength{\tabcolsep}{4pt}
\begin{lrbox}{\crtabbox}
\begin{tabularx}{\linewidth}{>{\raggedright\arraybackslash}Xrr}
\toprule
Quantity & \begin{tabular}[b]{@{}c@{}}ZINC-250k\\(full set)\end{tabular} & \begin{tabular}[b]{@{}c@{}}ZINC-250k,\\stereo stripped\end{tabular} \\
\midrule
Molecules encoded & $249{,}455$ & $249{,}455$ \\
Molecules with dropped tokens (count, share) & $157{,}337\;(63.1\,\%)$ & $5\;(0.0\,\%)$ \\
$\bar r$ = mean $\|\mu(x)\|$ $\pm$ SD & $7.9834 \pm 0.709$ & $7.9751 \pm 0.692$ \\
5th and 95th percentile of $\|\mu\|$ & $6.81;\ 9.14$ & $6.83;\ 9.11$ \\
$\mathrm{tr}\,\Sigma$ and $\|\bar\mu\|^2$ (two values) & $59.10;\ 5.13$ & $59.37;\ 4.71$ \\
Participation ratio $R_0=\mathrm{tr}(\Sigma)^2/\mathrm{tr}(\Sigma^2)$ & $64.7$ & $65.5$ \\
Principal components for 90\,\% and 99\,\% of the variance & $66;\ 92$ & $67;\ 93$ \\
Model radius $\sqrt{m}-\mathrm{Var}(\|\mu\|^2)/(8m^{3/2})$ (error vs $\bar r$, \%) & $7.9861\;(+0.03)$ & $7.9766\;(+0.02)$ \\
Model width $\sqrt{\mathrm{Var}(\|\mu\|^2)}/(2\sqrt{m})$ & $0.678$ & $0.676$ \\
Relative width SD$/\bar r$ vs $1/\sqrt{2R_0}$ & $0.0888 \text{ vs } 0.0879$ & $0.0868 \text{ vs } 0.0874$ \\
$\bar r/\sqrt{D-1/2}$ & $0.499$ & $0.499$ \\
\bottomrule
\end{tabularx}
\end{lrbox}
\ifdim\wd\crtabbox>\linewidth\resizebox{\linewidth}{!}{\usebox{\crtabbox}}\else\usebox{\crtabbox}\fi
\end{table}

%% file: tab_radius.tex
\ifdefined\crtabbox\else\newsavebox{\crtabbox}\fi
\begin{table}[t]
\centering
\caption{\textbf{Radius of \geowalk{} proposals.} Three blocks, each against its own reference row. Fixed radius: every proposal is placed at radius $r$. The default estimates $\bar r$ per run from the five cold-start latents. Radius estimator: the statistic of the five cold-start norms that sets $\bar r$. Thick shells: the radius of each proposal is drawn independently from the stated range. The bounds $p_{05}$ and $p_{95}$ are the 5th and 95th percentiles of the encoder-mean norms of ZINC-250k. All blocks use the same $n=440$ completed cells, as in Table~\ref{tab:main}. Means are means of per-task means. Bold marks the highest mean of each block. The difference $\Delta$ is the row minus the block's reference on the same cells. The effect size $d$ is Cohen's $d$ of the cell differences. The column $p_{\mathrm{BH}}$ is the two-sided Wilcoxon signed-rank test over the cells. It is Benjamini--Hochberg corrected within each block: 6 rows for the fixed radius, 2 for the estimator, 3 for the thick shells. Tasks counts the tasks on which the row has the higher mean. The column $p_{\mathrm{task}}$ is the exact two-sided Wilcoxon test on the 11 per-task mean differences, not corrected.}
\label{tab:radius}
\small
\setlength{\tabcolsep}{4pt}
\begin{lrbox}{\crtabbox}
\begin{tabularx}{\linewidth}{>{\raggedright\arraybackslash}Xrrrrrr}
\toprule
Variant & Mean & $\Delta$ & $d$ & $p_{\mathrm{BH}}$ & \begin{tabular}[b]{@{}c@{}}Tasks\\row\\higher\end{tabular} & $p_{\mathrm{task}}$ \\
\midrule
\multicolumn{7}{@{}l}{\textit{Fixed radius $r$ (reference $r=8$)}} \\
$r=4$ & $0.4124$ & $-0.0103$ & $-0.19$ & $0.009$ & 3/11 & $0.24$ \\
$r=6$ & $0.4187$ & $-0.0040$ & $-0.08$ & $0.20$ & 5/11 & $0.64$ \\
$r=10$ & $0.4204$ & $-0.0022$ & $-0.04$ & $0.12$ & 5/11 & $0.64$ \\
$r=12$ & $0.4199$ & $-0.0028$ & $-0.05$ & $0.064$ & 5/11 & $0.70$ \\
$r=16$ & $0.4117$ & $-0.0110$ & $-0.15$ & $<0.001$ & 3/11 & $0.37$ \\
per-run $\bar r$ (default) & $\mathbf{0.4226}$ & $0.0000$ & $0.00$ & $0.99$ & 5/11 & $1.00$ \\
\midrule
$r=8$ (reference) & $\mathbf{0.4227}$ & -- & -- & -- & -- & -- \\
\midrule
\multicolumn{7}{@{}l}{\textit{Radius estimator (reference: mean of the five norms)}} \\
median of the five cold-start norms & $0.4207$ & $-0.0019$ & $-0.07$ & $0.074$ & 2/11 & $0.083$ \\
fixed $7.98$ (ZINC mean norm) & $0.4223$ & $-0.0004$ & $-0.01$ & $0.82$ & 5/11 & $0.52$ \\
\midrule
mean of the five cold-start norms (reference) & $\mathbf{0.4226}$ & -- & -- & -- & -- & -- \\
\midrule
\multicolumn{7}{@{}l}{\textit{Thick shells (reference: single shell $\bar r$)}} \\
$r\sim\bar r\,U[0.9,1.1]$ & $0.4183$ & $-0.0043$ & $-0.08$ & $0.099$ & 3/11 & $0.054$ \\
$r\sim\bar r\,U[0.8,1.2]$ & $0.4195$ & $-0.0032$ & $-0.05$ & $0.13$ & 3/11 & $0.17$ \\
$r\sim U[p_{05},p_{95}]$ & $0.4204$ & $-0.0022$ & $-0.04$ & $0.13$ & 4/11 & $0.21$ \\
\midrule
single shell $\bar r$ (default) (reference) & $\mathbf{0.4226}$ & -- & -- & -- & -- & -- \\
\bottomrule
\end{tabularx}
\end{lrbox}
\ifdim\wd\crtabbox>\linewidth\resizebox{\linewidth}{!}{\usebox{\crtabbox}}\else\usebox{\crtabbox}\fi
\end{table}

%% file: tab_convex.tex
\ifdefined\crtabbox\else\newsavebox{\crtabbox}\fi
\begin{table}[t]
\centering
\caption{\textbf{Samplers inside the latent cloud, and direction controls.} First block: surrogate-free samplers against random search, which samples uniformly on the shell. The samplers draw convex combinations of $k$ latents with Dirichlet(1) weights, or draw from a Gaussian fitted to the latents, a sample of 3,000 encoded ZINC-250k molecules. $\overline{\|z\|}$ is the mean norm of the candidates. For random search it is $\bar r$, the radius of each run. Second block: two direction controls. Random search on the sphere in the span of the top 64 principal components is compared with random search on the full shell. The GP (DSP prior) with qLogEI argmax over cube directions rescaled to $\bar r$ is compared with the same GP on the uniform shell. All rows use the same $n=440$ completed cells, as in Table~\ref{tab:main}. Means are means of per-task means. Bold marks the highest mean of the first block. The difference $\Delta$ is the row minus its reference on the same cells. The effect size $d$ is Cohen's $d$ of the cell differences. The column $p_{\mathrm{BH}}$ is the two-sided Wilcoxon signed-rank test over the cells. It is Benjamini--Hochberg corrected within each block, over 8 rows and 2 comparisons. Tasks counts the tasks on which the row has the higher mean, of the 11. The column $p_{\mathrm{task}}$ is the exact two-sided Wilcoxon test on the 11 per-task mean differences, not corrected.}
\label{tab:convex}
\small
\setlength{\tabcolsep}{4pt}
\begin{lrbox}{\crtabbox}
\begin{tabularx}{\linewidth}{>{\raggedright\arraybackslash}Xrrrrrrr}
\toprule
Variant & Mean & $\overline{\|z\|}$ & $\Delta$ & $d$ & $p_{\mathrm{BH}}$ & \begin{tabular}[b]{@{}c@{}}Tasks\\row\\higher\end{tabular} & $p_{\mathrm{task}}$ \\
\midrule
convex, $k=1$ & $0.3838$ & $7.98$ & $-0.0280$ & $-0.31$ & $<0.001$ & 3/11 & $0.12$ \\
convex, $k=2$ & $0.3888$ & $6.62$ & $-0.0230$ & $-0.27$ & $<0.001$ & 4/11 & $0.24$ \\
convex, $k=3$ & $0.3909$ & $5.83$ & $-0.0209$ & $-0.24$ & $<0.001$ & 3/11 & $0.32$ \\
convex, $k=4$ & $0.3829$ & $5.32$ & $-0.0289$ & $-0.31$ & $<0.001$ & 3/11 & $0.17$ \\
convex, $k=8$ & $0.3893$ & $4.24$ & $-0.0225$ & $-0.23$ & $<0.001$ & 4/11 & $0.37$ \\
convex, $k=16$ & $0.3855$ & $3.46$ & $-0.0263$ & $-0.26$ & $<0.001$ & 3/11 & $0.32$ \\
convex, $k=64$ & $0.3823$ & $2.62$ & $-0.0295$ & $-0.27$ & $<0.001$ & 3/11 & $0.32$ \\
Gaussian fit $N(\hat\mu,\hat\Sigma)$ & $0.3840$ & $7.99$ & $-0.0278$ & $-0.33$ & $<0.001$ & 3/11 & $0.12$ \\
\midrule
uniform on the shell (random search) (reference) & $\mathbf{0.4118}$ & $\bar r$ & -- & -- & -- & -- & -- \\
\bottomrule
\end{tabularx}
\end{lrbox}
\ifdim\wd\crtabbox>\linewidth\resizebox{\linewidth}{!}{\usebox{\crtabbox}}\else\usebox{\crtabbox}\fi
\par\medskip
{\footnotesize\textit{Direction controls at the shell radius (reference: the same method with uniform directions on the shell)}}\par\smallskip
\begin{lrbox}{\crtabbox}
\begin{tabularx}{\linewidth}{>{\raggedright\arraybackslash}Xrrrrrrrr}
\toprule
Comparison & $n$ & Variant & Shell & $\Delta$ & $d$ & $p_{\mathrm{BH}}$ & \begin{tabular}[b]{@{}c@{}}Tasks\\variant\\higher\end{tabular} & $p_{\mathrm{task}}$ \\
\midrule
Random search: top-64 PCA sphere vs full sphere & 440 & $0.4069$ & $0.4118$ & $-0.0049$ & $-0.08$ & $0.046$ & 3/11 & $0.17$ \\
GP on the shell: cube directions rescaled to $\bar r$ vs uniform directions & 440 & $0.4220$ & $0.4228$ & $-0.0008$ & $-0.01$ & $0.91$ & 6/11 & $0.90$ \\
\bottomrule
\end{tabularx}
\end{lrbox}
\ifdim\wd\crtabbox>\linewidth\resizebox{\linewidth}{!}{\usebox{\crtabbox}}\else\usebox{\crtabbox}\fi
\end{table}

%% file: tab_factorial_contrasts.tex
\ifdefined\crtabbox\else\newsavebox{\crtabbox}\fi
\begin{table}[t]
\centering
\caption{\textbf{The five contrasts of the factorial experiment of Section~\ref{sec:boxsphere}.} Differences of the cross-means of Table~\ref{tab:factorial} on the same $n=440$ matched cells. $\Delta$ is the mean of per-task means, Tasks counts the tasks with a positive contrast, and $p_{\mathrm{task}}^{\mathrm{BH}}$ is the task-level test, corrected within the five contrasts; bold marks $p_{\mathrm{task}}^{\mathrm{BH}}<0.05$. The cell-level statistics and the bootstrap intervals are in the full table in the repository.}
\label{tab:factorial_contrasts}
\small
\begin{lrbox}{\crtabbox}
\begin{tabularx}{\linewidth}{>{\raggedright\arraybackslash}Xrrr}
\toprule
Contrast & $\Delta$ & \begin{tabular}[b]{@{}c@{}}Tasks\\$\Delta>0$\end{tabular} & $p_{\mathrm{task}}^{\mathrm{BH}}$ \\
\midrule
shell minus box, with candidate-set argmax & $+0.0303$ & 9/11 & $\mathbf{0.017}$ \\
shell minus box, with L-BFGS-B & $+0.0010$ & 6/11 & $0.90$ \\
candidate-set argmax minus L-BFGS-B, on the shell & $+0.0229$ & 9/11 & $\mathbf{0.017}$ \\
candidate-set argmax minus L-BFGS-B, in the box & $-0.0064$ & 1/11 & $\mathbf{0.012}$ \\
interaction: the first row minus the second & $+0.0294$ & 9/11 & $\mathbf{0.012}$ \\
\bottomrule
\end{tabularx}
\end{lrbox}
\ifdim\wd\crtabbox>\linewidth\resizebox{\linewidth}{!}{\usebox{\crtabbox}}\else\usebox{\crtabbox}\fi
\end{table}

%% file: checklist.tex
\section*{NeurIPS Paper Checklist}

\begin{enumerate}

\item {\bf Claims}
    \item[] Question: Do the main claims made in the abstract and introduction accurately reflect the paper's contributions and scope?
    \item[] Answer: \answerYes{}
    \item[] Justification: The claims are supported in Section~\ref{sec:experiments}: Table~\ref{tab:main} compares the methods on one matched set (its full version, with every row and the cell-level statistics, is in the repository), Table~\ref{tab:boxsphere} gives the three recipe rows, Table~\ref{tab:factorial} the factorial experiment whose contrasts (Table~\ref{tab:factorial_contrasts}, Appendix~\ref{app:contrasts}) carry the central claim, and Table~\ref{tab:crossdomain} compares \geowalk{} and \geocma{} across benchmarks; Appendices~\ref{app:algorithms}--\ref{app:contrasts} give the details. The primary test is the exact task-level Wilcoxon test over the 11 tasks, with Benjamini--Hochberg correction within families. The abstract and conclusion state that this is a controlled study of a Gaussian-prior VAE benchmark, not a recommendation to use VAEs for molecular optimization; Section~\ref{sec:setup} and the Limitations restrict it to $N=30$ and the 11 GuacaMol tasks of the protocol.

\item {\bf Limitations}
    \item[] Question: Does the paper discuss the limitations of the work performed by the authors?
    \item[] Answer: \answerYes{}
    \item[] Justification: The Limitations paragraph of Section~\ref{sec:discussion} states that the advantage of the shell is shown only for candidate sets drawn uniformly from the whole support; that the evidence comes only from the Gaussian-prior VAE benchmark studied and the 11 tasks of the protocol; that on nine further tasks the shell was ahead on three; that a second VAE (Chembed) gave no detected gain; that budgets above $N=30$ come from stored runs; and that flows, diffusion models and language-model spaces were not tested.

\item {\bf Theory assumptions and proofs}
    \item[] Question: For each theoretical result, does the paper provide the full set of assumptions and a complete (and correct) proof?
    \item[] Answer: \answerNA{}
    \item[] Justification: The paper is empirical and presents no theorems. The shell statistics of Section~\ref{sec:scaffold} and Appendix~\ref{app:radius} (mean norm, effective rank, a second-order description of the radius by two moments) are measured on the encoder means of the ZINC-250k reference set and describe the shell on the same encodings; they are not derived as a guarantee.

    \item {\bf Experimental result reproducibility}
    \item[] Question: Does the paper fully disclose all the information needed to reproduce the main experimental results of the paper to the extent that it affects the main claims and/or conclusions of the paper (regardless of whether the code and data are provided or not)?
    \item[] Answer: \answerYes{}
    \item[] Justification: Algorithms~\ref{alg:geowalk} and~\ref{alg:geocma} give complete pseudocode for \geowalk{} and \geocma{}, with their constants in Sections~\ref{sec:geowalk} and~\ref{sec:geocma}. Section~\ref{sec:setup} specifies the tasks, seeds, cold-start protocol, budget accounting and statistical protocol, and Section~\ref{sec:scaffold} defines the acquisition, both maximizers, the box and the GP constructor. Appendix~\ref{app:baselines} summarizes every baseline as it ran; the method cards in the repository give every deviation from the published method. The code, the run configurations, the result files and the generator of every table, figure and number are released (``Code and data availability'', Section~\ref{sec:discussion}). The SELFIES VAE weights are the public release accompanying LOL-BO~\citep{maus2022lolbo}.

\item {\bf Open access to data and code}
    \item[] Question: Does the paper provide open access to the data and code, with sufficient instructions to faithfully reproduce the main experimental results, as described in supplemental material?
    \item[] Answer: \answerYes{}
    \item[] Justification: The code, the run configurations and the result files are released in the public repository given in the ``Code and data availability'' paragraph of Section~\ref{sec:discussion}, with instructions and the generator that writes every table, figure and number from the result files. GuacaMol, LassoBench, NAS-Bench-101 and the SELFIES VAE released with LOL-BO are public. ZINC-250k is the public subset distributed by~\citet{gomez2018automatic}; it serves here as a reference set, not as the VAE's training set.

\item {\bf Experimental setting/details}
    \item[] Question: Does the paper specify all the training and test details (e.g., data splits, hyperparameters, how they were chosen, type of optimizer) necessary to understand the results?
    \item[] Answer: \answerYes{}
    \item[] Justification: Section~\ref{sec:setup} specifies the tasks, seeds, cold-start protocol and statistical procedure, and Section~\ref{sec:scaffold} the acquisition, the maximizers, the box and the GP. Sections~\ref{sec:geowalk} and~\ref{sec:geocma} and Table~\ref{tab:hyper} give the settings of \geowalk{} and \geocma{}, and Appendix~\ref{app:sensitivity} their sensitivity. Baselines are adaptations of the published methods run in our harness, summarized in Appendix~\ref{app:baselines}, with every known deviation listed in the method cards of the repository; the factorial experiment is described in Section~\ref{sec:boxsphere}.

\item {\bf Experiment statistical significance}
    \item[] Question: Does the paper report error bars suitably and correctly defined or other appropriate information about the statistical significance of the experiments?
    \item[] Answer: \answerYes{}
    \item[] Justification: The primary test of every comparison is the exact two-sided Wilcoxon signed-rank test on the 11 per-task mean differences, reported with the number of tasks on which a method is ahead (Section~\ref{sec:setup}). Benjamini--Hochberg correction is applied within families: the five factorial contrasts, the three recipe rows of Table~\ref{tab:boxsphere}, the published-recipe tests of Table~\ref{tab:published}, and the LassoBench tests of the published methods; the task-level $p$ of Tables~\ref{tab:main} and~\ref{tab:crossdomain} is uncorrected. The full version of Table~\ref{tab:factorial_contrasts} in the repository also reports a percentile bootstrap interval over tasks for each contrast. Cell-level Wilcoxon tests and Cohen's $d$ are secondary, and every table prints the size of its matched set. Bands in Figure~\ref{fig:convergence} are standard errors over seeds.

\item {\bf Experiments compute resources}
    \item[] Question: For each experiment, does the paper provide sufficient information on the computer resources (type of compute workers, memory, time of execution) needed to reproduce the experiments?
    \item[] Answer: \answerYes{}
    \item[] Justification: Appendix~\ref{app:compute} names the machines and gives the total process-hours; a table in the repository gives the process-hours and the median time of one run for every block of runs, recomputed from the times recorded in the result files by a script of the repository. The runs other than the stored runs took about \crComputeGpuServerHours{} process-hours on a server with two NVIDIA L40S GPUs and about \crComputeCpuServerHours{} on the CPUs of a second server, most of them for checks of the published recipes. The stored runs that printed tables read took about \crComputeStoredPrintedHours{} process-hours. The project also used compute for preliminary and failed runs that the paper does not report.

\item {\bf Code of ethics}
    \item[] Question: Does the research conducted in the paper conform, in every respect, with the NeurIPS Code of Ethics \url{https://neurips.cc/public/EthicsGuidelines}?
    \item[] Answer: \answerYes{}
    \item[] Justification: The research uses publicly available benchmarks (GuacaMol, LassoBench, NAS-Bench-101) and does not involve human subjects, private data, or personally identifiable information. Dual-use risks of molecular optimization are acknowledged in Section~\ref{sec:discussion}.

\item {\bf Broader impacts}
    \item[] Question: Does the paper discuss both potential positive societal impacts and negative societal impacts of the work performed?
    \item[] Answer: \answerYes{}
    \item[] Justification: The Limitations paragraph of Section~\ref{sec:discussion} notes that better low-budget molecular optimization could lower experimental cost (positive) and that it has dual-use risks if applied to harmful objectives (negative).

\item {\bf Safeguards}
    \item[] Question: Does the paper describe safeguards that have been put in place for responsible release of data or models that have a high risk for misuse (e.g., pre-trained language models, image generators, or scraped datasets)?
    \item[] Answer: \answerNA{}
    \item[] Justification: The paper releases optimization code, run configurations and result files only; the underlying SELFIES VAE and oracle implementations are pre-existing public assets released by their original authors. No new generative model, pre-trained language model, or large-scale scraped dataset is released.

\item {\bf Licenses for existing assets}
    \item[] Question: Are the creators or original owners of assets (e.g., code, data, models), used in the paper, properly credited and are the license and terms of use explicitly mentioned and properly respected?
    \item[] Answer: \answerYes{}
    \item[] Justification: Every external asset is cited, and Appendix~\ref{app:compute} gives its license and version. Our port of the official TuRBO code~\citep{eriksson2019turbo} (\url{https://github.com/uber-research/TuRBO}) is distributed under the non-commercial TuRBO license, with the change notice it requires. Third-party files keep their own licenses, listed in the repository's third-party notices.

\item {\bf New assets}
    \item[] Question: Are new assets introduced in the paper well documented and is the documentation provided alongside the assets?
    \item[] Answer: \answerYes{}
    \item[] Justification: \geowalk{}, \geocma{}, the experiment scripts, the run configurations and the result files are released in the public repository given in the ``Code and data availability'' paragraph of Section~\ref{sec:discussion}. It contains installation instructions, the generator that reproduces every table, figure and number of the paper, the full card and the run status of every method, a license file (MIT for the authors' code), and the notices of the third-party and ported files under their own licenses.

\item {\bf Crowdsourcing and research with human subjects}
    \item[] Question: For crowdsourcing experiments and research with human subjects, does the paper include the full text of instructions given to participants and screenshots, if applicable, as well as details about compensation (if any)?
    \item[] Answer: \answerNA{}
    \item[] Justification: This research does not involve crowdsourcing or human subjects.

\item {\bf Institutional review board (IRB) approvals or equivalent for research with human subjects}
    \item[] Question: Does the paper describe potential risks incurred by study participants, whether such risks were disclosed to the subjects, and whether Institutional Review Board (IRB) approvals (or an equivalent approval/review based on the requirements of your country or institution) were obtained?
    \item[] Answer: \answerNA{}
    \item[] Justification: This research does not involve human subjects.

\item {\bf Declaration of LLM usage}
    \item[] Question: Does the paper describe the usage of LLMs if it is an important, original, or non-standard component of the core methods in this research? Note that if the LLM is used only for writing, editing, or formatting purposes and does \emph{not} impact the core methodology, scientific rigor, or originality of the research, declaration is not required.
    \item[] Answer: \answerNA{}
    \item[] Justification: The proposed methods (\geowalk{} and \geocma{}) contain no LLM component. As declared on OpenReview, LLMs were used for data processing and filtering, implementing standard methods, editing, and drafting sections of the paper; this does not affect the core methodology.

\end{enumerate}

%% file: main.bbl
\begin{thebibliography}{44}
\providecommand{\natexlab}[1]{#1}
\providecommand{\url}[1]{\texttt{#1}}
\expandafter\ifx\csname urlstyle\endcsname\relax
  \providecommand{\doi}[1]{doi: #1}\else
  \providecommand{\doi}{doi: \begingroup \urlstyle{rm}\Url}\fi

\bibitem[Eriksson et~al.(2019)Eriksson, Pearce, Gardner, Turner, and
  Poloczek]{eriksson2019turbo}
David Eriksson, Michael Pearce, Jacob~R. Gardner, Ryan~D. Turner, and Matthias
  Poloczek.
\newblock Scalable global optimization via local {B}ayesian optimization.
\newblock In \emph{Advances in Neural Information Processing Systems},
  volume~32. Curran Associates, Inc., 2019.
\newblock URL
  \url{https://proceedings.neurips.cc/paper_files/paper/2019/hash/6c990b7aca7bc7058f5e98ea909e924b-Abstract.html}.

\bibitem[Maus et~al.(2022)Maus, Jones, Moore, Kusner, Bradshaw, and
  Gardner]{maus2022lolbo}
Natalie Maus, Haydn~T. Jones, Juston~S. Moore, Matt~J. Kusner, John Bradshaw,
  and Jacob~R. Gardner.
\newblock Local latent space {B}ayesian optimization over structured inputs.
\newblock In \emph{Advances in Neural Information Processing Systems},
  volume~35, pages 34505--34518. Curran Associates, Inc., 2022.
\newblock \doi{10.52202/068431-2500}.

\bibitem[Chu et~al.(2024)Chu, Park, Lee, and Kim]{chu2024invbo}
Jaewon Chu, Jinyoung Park, Seunghun Lee, and Hyunwoo~J. Kim.
\newblock Inversion-based latent {B}ayesian optimization.
\newblock In \emph{Advances in Neural Information Processing Systems},
  volume~37, pages 68258--68286. Curran Associates, Inc., 2024.
\newblock \doi{10.52202/079017-2180}.

\bibitem[Krenn et~al.(2020)Krenn, H{\"a}se, Nigam, Friederich, and
  Aspuru-Guzik]{krenn2020selfies}
Mario Krenn, Florian H{\"a}se, AkshatKumar Nigam, Pascal Friederich, and Alan
  Aspuru-Guzik.
\newblock Self-referencing embedded strings ({SELFIES}): A 100\% robust
  molecular string representation.
\newblock \emph{Machine Learning: Science and Technology}, 1\penalty0
  (4):\penalty0 045024, 2020.
\newblock \doi{10.1088/2632-2153/aba947}.

\bibitem[Brown et~al.(2019)Brown, Fiscato, Segler, and
  Vaucher]{brown2019guacamol}
Nathan Brown, Marco Fiscato, Marwin H.~S. Segler, and Alain~C. Vaucher.
\newblock {GuacaMol}: Benchmarking models for de novo molecular design.
\newblock \emph{Journal of Chemical Information and Modeling}, 59\penalty0
  (3):\penalty0 1096--1108, 2019.
\newblock \doi{10.1021/acs.jcim.8b00839}.

\bibitem[Irwin et~al.(2012)Irwin, Sterling, Mysinger, Bolstad, and
  Coleman]{irwin2012zinc}
John~J. Irwin, Teague Sterling, Michael~M. Mysinger, Erin~S. Bolstad, and
  Ryan~G. Coleman.
\newblock {ZINC}: A free tool to discover chemistry for biology.
\newblock \emph{Journal of Chemical Information and Modeling}, 52\penalty0
  (7):\penalty0 1757--1768, 2012.
\newblock \doi{10.1021/ci3001277}.

\bibitem[G{\'o}mez-Bombarelli et~al.(2018)G{\'o}mez-Bombarelli, Wei, Duvenaud,
  Hern{\'a}ndez-Lobato, S{\'a}nchez-Lengeling, Sheberla, Aguilera-Iparraguirre,
  Hirzel, Adams, and Aspuru-Guzik]{gomez2018automatic}
Rafael G{\'o}mez-Bombarelli, Jennifer~N. Wei, David Duvenaud, Jos{\'e}~Miguel
  Hern{\'a}ndez-Lobato, Benjam{\'\i}n S{\'a}nchez-Lengeling, Dennis Sheberla,
  Jorge Aguilera-Iparraguirre, Timothy~D. Hirzel, Ryan~P. Adams, and Al{\'a}n
  Aspuru-Guzik.
\newblock Automatic chemical design using a data-driven continuous
  representation of molecules.
\newblock \emph{ACS Central Science}, 4\penalty0 (2):\penalty0 268--276, 2018.
\newblock \doi{10.1021/acscentsci.7b00572}.

\bibitem[{\v{S}}ehi{\'c} et~al.(2022){\v{S}}ehi{\'c}, Gramfort, Salmon, and
  Nardi]{sehic2022lassobench}
Kenan {\v{S}}ehi{\'c}, Alexandre Gramfort, Joseph Salmon, and Luigi Nardi.
\newblock {LassoBench}: A high-dimensional hyperparameter optimization
  benchmark suite for {L}asso.
\newblock In \emph{Proceedings of the First International Conference on
  Automated Machine Learning}, volume 188 of \emph{Proceedings of Machine
  Learning Research}, pages 2/1--24. PMLR, 2022.
\newblock URL \url{https://proceedings.mlr.press/v188/sehic22a.html}.

\bibitem[Ying et~al.(2019)Ying, Klein, Christiansen, Real, Murphy, and
  Hutter]{ying2019nasbench}
Chris Ying, Aaron Klein, Eric Christiansen, Esteban Real, Kevin Murphy, and
  Frank Hutter.
\newblock {NAS-Bench-101}: Towards reproducible neural architecture search.
\newblock In \emph{Proceedings of the 36th International Conference on Machine
  Learning}, volume~97 of \emph{Proceedings of Machine Learning Research},
  pages 7105--7114. PMLR, 2019.
\newblock URL \url{https://proceedings.mlr.press/v97/ying19a.html}.

\bibitem[Lee et~al.(2023)Lee, Chu, Kim, Ko, and Kim]{lee2023cobo}
Seunghun Lee, Jaewon Chu, Sihyeon Kim, Juyeon Ko, and Hyunwoo~J. Kim.
\newblock Advancing {B}ayesian optimization via learning correlated latent
  space.
\newblock In \emph{Advances in Neural Information Processing Systems},
  volume~36, pages 48906--48917. Curran Associates, Inc., 2023.
\newblock \doi{10.52202/075280-2124}.

\bibitem[Lee et~al.(2025)Lee, Park, Chu, Yoon, and Kim]{lee2025nfbo}
Seunghun Lee, Jinyoung Park, Jaewon Chu, Minseo Yoon, and Hyunwoo~J. Kim.
\newblock Latent {B}ayesian optimization via autoregressive normalizing flows.
\newblock In \emph{The Thirteenth International Conference on Learning
  Representations}, 2025.
\newblock URL \url{https://openreview.net/forum?id=ZCOwwRAaEl}.

\bibitem[Maus et~al.(2024)Maus, Kim, Pleiss, Eriksson, Cunningham, and
  Gardner]{maus2024eulbo}
Natalie Maus, Kyurae Kim, Geoff Pleiss, David Eriksson, John~P. Cunningham, and
  Jacob~R. Gardner.
\newblock Approximation-aware {B}ayesian optimization.
\newblock In \emph{Advances in Neural Information Processing Systems},
  volume~37, pages 21114--21140. Curran Associates, Inc., 2024.
\newblock \doi{10.52202/079017-0665}.

\bibitem[Moss et~al.(2025)Moss, Ober, and Diethe]{moss2025cowboys}
Henry Moss, Sebastian~W. Ober, and Tom Diethe.
\newblock Return of the latent space {COWBOYS}: Re-thinking the use of {VAE}s
  for {B}ayesian optimisation of structured spaces.
\newblock In \emph{Proceedings of the 42nd International Conference on Machine
  Learning}, volume 267 of \emph{Proceedings of Machine Learning Research},
  pages 44956--44970. PMLR, 2025.
\newblock URL \url{https://proceedings.mlr.press/v267/moss25a.html}.

\bibitem[Papenmeier et~al.(2022)Papenmeier, Nardi, and
  Poloczek]{papenmeier2022baxus}
Leonard Papenmeier, Luigi Nardi, and Matthias Poloczek.
\newblock Increasing the scope as you learn: Adaptive {B}ayesian optimization
  in nested subspaces.
\newblock In \emph{Advances in Neural Information Processing Systems},
  volume~35, pages 11586--11601. Curran Associates, Inc., 2022.
\newblock \doi{10.52202/068431-0842}.

\bibitem[Eriksson and Jankowiak(2021)]{eriksson2021saasbo}
David Eriksson and Martin Jankowiak.
\newblock High-dimensional {B}ayesian optimization with sparse axis-aligned
  subspaces.
\newblock In \emph{Proceedings of the Thirty-Seventh Conference on Uncertainty
  in Artificial Intelligence}, volume 161 of \emph{Proceedings of Machine
  Learning Research}, pages 493--503. PMLR, 2021.
\newblock URL \url{https://proceedings.mlr.press/v161/eriksson21a.html}.

\bibitem[Wang et~al.(2016)Wang, Hutter, Zoghi, Matheson, and
  de~Freitas]{wang2016rembo}
Ziyu Wang, Frank Hutter, Masrour Zoghi, David Matheson, and Nando de~Freitas.
\newblock {B}ayesian optimization in a billion dimensions via random
  embeddings.
\newblock \emph{Journal of Artificial Intelligence Research}, 55:\penalty0
  361--387, 2016.
\newblock \doi{10.1613/jair.4806}.

\bibitem[Nayebi et~al.(2019)Nayebi, Munteanu, and Poloczek]{nayebi2019hesbo}
Amin Nayebi, Alexander Munteanu, and Matthias Poloczek.
\newblock A framework for {B}ayesian optimization in embedded subspaces.
\newblock In \emph{Proceedings of the 36th International Conference on Machine
  Learning}, volume~97 of \emph{Proceedings of Machine Learning Research},
  pages 4752--4761. PMLR, 2019.
\newblock URL \url{https://proceedings.mlr.press/v97/nayebi19a.html}.

\bibitem[Letham et~al.(2020)Letham, Calandra, Rai, and Bakshy]{letham2020alebo}
Ben Letham, Roberto Calandra, Akshara Rai, and Eytan Bakshy.
\newblock Re-examining linear embeddings for high-dimensional {B}ayesian
  optimization.
\newblock In \emph{Advances in Neural Information Processing Systems},
  volume~33, pages 1546--1558. Curran Associates, Inc., 2020.
\newblock URL
  \url{https://proceedings.neurips.cc/paper_files/paper/2020/hash/10fb6cfa4c990d2bad5ddef4f70e8ba2-Abstract.html}.

\bibitem[Hansen and Ostermeier(2001)]{hansen2001cmaes}
Nikolaus Hansen and Andreas Ostermeier.
\newblock Completely derandomized self-adaptation in evolution strategies.
\newblock \emph{Evolutionary Computation}, 9\penalty0 (2):\penalty0 159--195,
  2001.
\newblock \doi{10.1162/106365601750190398}.

\bibitem[Hansen et~al.(2019)Hansen, Akimoto, and Baudis]{hansen2019pycma}
Nikolaus Hansen, Youhei Akimoto, and Petr Baudis.
\newblock {CMA-ES/pycma} on {G}ithub.
\newblock Zenodo, 2019.
\newblock URL \url{https://doi.org/10.5281/zenodo.2559634}.

\bibitem[Doumont et~al.(2026)Doumont, Fan, Maus, Gardner, Moss, and
  Pleiss]{doumont2026still}
Colin Doumont, Donney Fan, Natalie Maus, Jacob~R. Gardner, Henry Moss, and
  Geoff Pleiss.
\newblock We still don't understand high-dimensional {B}ayesian optimization.
\newblock In \emph{Proceedings of The 29th International Conference on
  Artificial Intelligence and Statistics}, volume 300 of \emph{Proceedings of
  Machine Learning Research}, pages 2629--2637. PMLR, 2026.
\newblock URL \url{https://proceedings.mlr.press/v300/doumont26a.html}.

\bibitem[Jaquier et~al.(2020)Jaquier, Rozo, Calinon, and
  B{\"u}rger]{jaquier2020gabo}
No{\'e}mie Jaquier, Leonel Rozo, Sylvain Calinon, and Mathias B{\"u}rger.
\newblock {B}ayesian optimization meets {R}iemannian manifolds in robot
  learning.
\newblock In \emph{Proceedings of the Conference on Robot Learning}, volume 100
  of \emph{Proceedings of Machine Learning Research}, pages 233--246. PMLR,
  2020.
\newblock URL \url{https://proceedings.mlr.press/v100/jaquier20a.html}.

\bibitem[Jaquier et~al.(2022)Jaquier, Borovitskiy, Smolensky, Terenin, Asfour,
  and Rozo]{jaquier2022geometry}
No{\'e}mie Jaquier, Viacheslav Borovitskiy, Andrei Smolensky, Alexander
  Terenin, Tamim Asfour, and Leonel Rozo.
\newblock Geometry-aware {B}ayesian optimization in robotics using {R}iemannian
  {M}at{\'e}rn kernels.
\newblock In \emph{Proceedings of the 5th Conference on Robot Learning}, volume
  164 of \emph{Proceedings of Machine Learning Research}, pages 794--805. PMLR,
  2022.
\newblock URL \url{https://proceedings.mlr.press/v164/jaquier22a.html}.

\bibitem[Borovitskiy et~al.(2020)Borovitskiy, Terenin, Mostowsky, and
  Deisenroth]{borovitskiy2020matern}
Viacheslav Borovitskiy, Alexander Terenin, Peter Mostowsky, and Marc~Peter
  Deisenroth.
\newblock {M}at{\'e}rn {G}aussian processes on {R}iemannian manifolds.
\newblock In \emph{Advances in Neural Information Processing Systems},
  volume~33, pages 12426--12437. Curran Associates, Inc., 2020.
\newblock URL
  \url{https://proceedings.neurips.cc/paper_files/paper/2020/hash/92bf5e6240737e0326ea59846a83e076-Abstract.html}.

\bibitem[Hvarfner et~al.(2024)Hvarfner, Hellsten, and
  Nardi]{hvarfner2024vanilla}
Carl Hvarfner, Erik~Orm Hellsten, and Luigi Nardi.
\newblock Vanilla {B}ayesian optimization performs great in high dimensions.
\newblock In \emph{Proceedings of the 41st International Conference on Machine
  Learning}, volume 235 of \emph{Proceedings of Machine Learning Research},
  pages 20793--20817. PMLR, 2024.
\newblock URL \url{https://proceedings.mlr.press/v235/hvarfner24a.html}.

\bibitem[Papenmeier et~al.(2025)Papenmeier, Poloczek, and
  Nardi]{papenmeier2025understanding}
Leonard Papenmeier, Matthias Poloczek, and Luigi Nardi.
\newblock Understanding high-dimensional {B}ayesian optimization.
\newblock In \emph{Proceedings of the 42nd International Conference on Machine
  Learning}, volume 267 of \emph{Proceedings of Machine Learning Research},
  pages 47902--47923. PMLR, 2025.
\newblock URL \url{https://proceedings.mlr.press/v267/papenmeier25a.html}.

\bibitem[Xu et~al.(2025)Xu, Wang, Phillips, and Zhe]{xu2025standardgp}
Zhitong Xu, Haitao Wang, Jeff~M. Phillips, and Shandian Zhe.
\newblock Standard {G}aussian process is all you need for high-dimensional
  {B}ayesian optimization.
\newblock In \emph{The Thirteenth International Conference on Learning
  Representations}, 2025.
\newblock URL \url{https://openreview.net/forum?id=kX8h23UG6v}.

\bibitem[Tripp and Hern{\'a}ndez-Lobato(2023)]{tripp2023genetic}
Austin Tripp and Jos{\'e}~Miguel Hern{\'a}ndez-Lobato.
\newblock Genetic algorithms are strong baselines for molecule generation.
\newblock arXiv preprint, 2023.
\newblock URL \url{https://arxiv.org/abs/2310.09267}.

\bibitem[Gonz{\'a}lez-Duque et~al.(2024)Gonz{\'a}lez-Duque, Michael, Bartels,
  Zainchkovskyy, Hauberg, and Boomsma]{gonzalezduque2024hdbo_survey}
Miguel Gonz{\'a}lez-Duque, Richard Michael, Simon Bartels, Yevgen
  Zainchkovskyy, S{\o}ren Hauberg, and Wouter Boomsma.
\newblock A survey and benchmark of high-dimensional {B}ayesian optimization of
  discrete sequences.
\newblock In \emph{Advances in Neural Information Processing Systems},
  volume~37, pages 140478--140508. Curran Associates, Inc., 2024.
\newblock \doi{10.52202/079017-4459}.

\bibitem[Maus(2022)]{maus2022lolbocode}
Natalie Maus.
\newblock Local latent {B}ayesian optimization ({LOLBO}).
\newblock GitHub repository, official implementation of Local Latent Space
  Bayesian Optimization over Structured Inputs, 2022.
\newblock URL \url{https://github.com/nataliemaus/lolbo}.
\newblock Commit 0876147, 31 October 2022.

\bibitem[Wilcoxon(1945)]{wilcoxon1945individual}
Frank Wilcoxon.
\newblock Individual comparisons by ranking methods.
\newblock \emph{Biometrics Bulletin}, 1\penalty0 (6):\penalty0 80--83, 1945.
\newblock \doi{10.2307/3001968}.

\bibitem[Benjamini and Hochberg(1995)]{benjamini1995fdr}
Yoav Benjamini and Yosef Hochberg.
\newblock Controlling the false discovery rate: A practical and powerful
  approach to multiple testing.
\newblock \emph{Journal of the Royal Statistical Society: Series B
  (Methodological)}, 57\penalty0 (1):\penalty0 289--300, 1995.
\newblock \doi{10.1111/j.2517-6161.1995.tb02031.x}.

\bibitem[Arvanitidis et~al.(2018)Arvanitidis, Hansen, and
  Hauberg]{arvanitidis2018latent}
Georgios Arvanitidis, Lars~Kai Hansen, and S{\o}ren Hauberg.
\newblock Latent space oddity: On the curvature of deep generative models.
\newblock In \emph{International Conference on Learning Representations}, 2018.
\newblock URL \url{https://openreview.net/forum?id=SJzRZ-WCZ}.

\bibitem[Davidson et~al.(2018)Davidson, Falorsi, De~Cao, Kipf, and
  Tomczak]{davidson2018hyperspherical}
Tim~R. Davidson, Luca Falorsi, Nicola De~Cao, Thomas Kipf, and Jakub~M.
  Tomczak.
\newblock Hyperspherical variational auto-encoders.
\newblock In \emph{Proceedings of the Thirty-Fourth Conference on Uncertainty
  in Artificial Intelligence}, pages 856--865. AUAI Press, 2018.
\newblock URL \url{https://www.auai.org/uai2018/proceedings/papers/309.pdf}.

\bibitem[Xu and Durrett(2018)]{xu2018spherical}
Jiacheng Xu and Greg Durrett.
\newblock Spherical latent spaces for stable variational autoencoders.
\newblock In \emph{Proceedings of the 2018 Conference on Empirical Methods in
  Natural Language Processing}, pages 4503--4513. Association for Computational
  Linguistics, 2018.
\newblock \doi{10.18653/v1/D18-1480}.

\bibitem[De~Cao and Aziz(2020)]{decao2020powerspherical}
Nicola De~Cao and Wilker Aziz.
\newblock The {P}ower {S}pherical distribution.
\newblock In \emph{ICML 2020 Workshop on Invertible Neural Networks,
  Normalizing Flows, and Explicit Likelihood Models (INNF+)}, 2020.
\newblock URL \url{https://arxiv.org/abs/2006.04437}.

\bibitem[Balandat et~al.(2020)Balandat, Karrer, Jiang, Daulton, Letham, Wilson,
  and Bakshy]{balandat2020botorch}
Maximilian Balandat, Brian Karrer, Daniel~R. Jiang, Samuel Daulton, Ben Letham,
  Andrew~Gordon Wilson, and Eytan Bakshy.
\newblock {BoTorch}: A framework for efficient {M}onte-{C}arlo {B}ayesian
  optimization.
\newblock In \emph{Advances in Neural Information Processing Systems},
  volume~33, pages 21524--21538. Curran Associates, Inc., 2020.
\newblock URL
  \url{https://proceedings.neurips.cc/paper_files/paper/2020/hash/f5b1b89d98b7286673128a5fb112cb9a-Abstract.html}.

\bibitem[Rechenberg(1973)]{rechenberg1973}
Ingo Rechenberg.
\newblock \emph{Evolutionsstrategie: Optimierung technischer {S}ysteme nach
  {P}rinzipien der biologischen {E}volution}.
\newblock Number~15 in Problemata. Frommann-Holzboog, Stuttgart, 1973.
\newblock ISBN 3772803733.

\bibitem[J{\"a}gersk{\"u}pper(2007)]{jagerskuepper2007ppsn}
Jens J{\"a}gersk{\"u}pper.
\newblock Algorithmic analysis of a basic evolutionary algorithm for continuous
  optimization.
\newblock \emph{Theoretical Computer Science}, 379\penalty0 (3):\penalty0
  329--347, 2007.
\newblock \doi{10.1016/j.tcs.2007.02.042}.

\bibitem[Yan et~al.(2020)Yan, Zheng, Ao, Zeng, and Zhang]{yan2020arch2vec}
Shen Yan, Yu~Zheng, Wei Ao, Xiao Zeng, and Mi~Zhang.
\newblock Does unsupervised architecture representation learning help neural
  architecture search?
\newblock In \emph{Advances in Neural Information Processing Systems},
  volume~33, pages 12486--12498. Curran Associates, Inc., 2020.
\newblock URL
  \url{https://proceedings.neurips.cc/paper_files/paper/2020/hash/937936029af671cf479fa893db91cbdd-Abstract.html}.

\bibitem[Talibart and Gilis(2026)]{talibart2026chembed}
Hugo Talibart and Dimitri Gilis.
\newblock Learning a chemistry-aware latent space for molecular encoding and
  generation with a large-scale transformer variational autoencoder.
\newblock \emph{Journal of Cheminformatics}, 18\penalty0 (1):\penalty0 123,
  2026.
\newblock \doi{10.1186/s13321-026-01243-0}.

\bibitem[Paszke et~al.(2019)Paszke, Gross, Massa, Lerer, Bradbury, Chanan,
  Killeen, Lin, Gimelshein, Antiga, Desmaison, K{\"o}pf, Yang, DeVito, Raison,
  Tejani, Chilamkurthy, Steiner, Fang, Bai, and Chintala]{paszke2019pytorch}
Adam Paszke, Sam Gross, Francisco Massa, Adam Lerer, James Bradbury, Gregory
  Chanan, Trevor Killeen, Zeming Lin, Natalia Gimelshein, Luca Antiga, Alban
  Desmaison, Andreas K{\"o}pf, Edward Yang, Zachary DeVito, Martin Raison,
  Alykhan Tejani, Sasank Chilamkurthy, Benoit Steiner, Lu~Fang, Junjie Bai, and
  Soumith Chintala.
\newblock {PyTorch}: An imperative style, high-performance deep learning
  library.
\newblock In \emph{Advances in Neural Information Processing Systems},
  volume~32. Curran Associates, Inc., 2019.
\newblock URL
  \url{https://proceedings.neurips.cc/paper_files/paper/2019/hash/bdbca288fee7f92f2bfa9f7012727740-Abstract.html}.

\bibitem[rdk(2026)]{rdkitsoftware}
{RDKit}: Open-source cheminformatics.
\newblock Release 2026.03.1, \url{https://doi.org/10.5281/zenodo.19250388},
  2026.
\newblock URL \url{https://www.rdkit.org}.

\bibitem[Gardner et~al.(2018)Gardner, Pleiss, Weinberger, Bindel, and
  Wilson]{gardner2018gpytorch}
Jacob~R. Gardner, Geoff Pleiss, Kilian~Q. Weinberger, David Bindel, and
  Andrew~Gordon Wilson.
\newblock {GPyTorch}: Blackbox matrix-matrix {G}aussian process inference with
  {GPU} acceleration.
\newblock In \emph{Advances in Neural Information Processing Systems},
  volume~31. Curran Associates, Inc., 2018.
\newblock URL
  \url{https://proceedings.neurips.cc/paper_files/paper/2018/hash/27e8e17134dd7083b050476733207ea1-Abstract.html}.

\end{thebibliography}
